\pdfoutput=1
\documentclass[11pt]{article}

\usepackage{EMNLP2022}

\usepackage{times}
\usepackage{latexsym}

\usepackage[T1]{fontenc}
\usepackage[utf8]{inputenc}

\usepackage{microtype}

\usepackage{inconsolata}

\usepackage{amssymb,amsmath,amsthm}
\usepackage{mathtools}
\usepackage{stmaryrd}
\usepackage{hyperref}
\usepackage{bbm}
\usepackage{natbib}
\usepackage{csquotes}
\usepackage{caption}
\usepackage{subcaption}

\hypersetup{hidelinks}

\DeclarePairedDelimiter\abs{\lvert}{\rvert}%
\DeclarePairedDelimiter\norm{\lVert}{\rVert}%

\makeatletter
\let\oldabs\abs
\def\abs{\@ifstar{\oldabs}{\oldabs*}}
\let\oldnorm\norm
\def\norm{\@ifstar{\oldnorm}{\oldnorm*}}
\makeatother
\DeclarePairedDelimiter{\den}{\llbracket}{\rrbracket}

\DeclareMathOperator*{\EX}{\mathbb{E}}% expected value
\usepackage{xpatch}
\makeatletter
\AtBeginDocument{\xpatchcmd{\@thm}{\thm@headpunct{.}}{\thm@headpunct{}}{}{}}
\makeatother

\newtheorem{theorem}{Theorem}
\newtheorem{lemma}{Lemma}
\newtheorem{corollary}{Corollary}[theorem]
\newtheorem{corollarylem}{Corollary}[lemma]

\theoremstyle{definition}
\newtheorem{definition}{Definition}

\def\Snospace~{\S{}}

\usepackage{xcolor}

\usepackage{tikz}
\usetikzlibrary{shapes,decorations,arrows,calc,arrows.meta,fit,positioning,patterns}
\tikzset{
    -Latex,auto,node distance =1 cm and 1 cm,semithick,
    state/.style ={ellipse, draw, minimum width = 0.7 cm},
    point/.style = {circle, draw, inner sep=0.04cm,fill,node contents={}},
    bidirected/.style={Latex-Latex,dashed},
    el/.style = {inner sep=2pt, align=left, sloped}
}

\usepackage{centernot}

\newcommand{\notiff}{%
  \mathrel{{\ooalign{\hidewidth$\not\phantom{"}$\hidewidth\cr$\iff$}}}}
  
\renewcommand{\omega}{\$}

\usepackage{thm-restate}

\usepackage{cancel}
\usepackage{paralist}
\usepackage{xcolor}

\newcommand{\erratumcolor}{olive}

\title{Entailment Semantics Can Be Extracted from an Ideal Language Model}

\author{William Merrill \\
        New York University \\
        \And
        Alex Warstadt \\
        ETH Z\"urich \\
        \texttt{\{willm,linzen\}@nyu.edu alexanderscott.warstadt@inf.ethz.ch}
        \And
        Tal Linzen \\
        New York University \\
        }

\usepackage{tcolorbox}
\usepackage{graphicx}

\begin{document}

\color{\erratumcolor}

\section*{Erratum Note}

On October 6, 2023, Benjamin Spector and Emmanuel Chemla found a mistake in the main result (\autoref{thm:informative}) of this paper.\footnote{\color{\erratumcolor} The authors thank Sophie Hao, Noah A. Smith, and Zhaofeng Wu for feedback on this erratum.} The technical problem leads to an alternate interpretation of our results. Our distributional ``entailment test'' does not actually detect entailment, but, rather, detects either entailment or \emph{near contradiction}, meaning two sentences are only consistent in a rare set of worlds, without distinguishing which.

\subsection*{Implications}
We believe near contradiction may be rare compared to entailment, so the test may still tend to identify entailment correctly assuming realistic data distributions. However, our paper's main goal was never to propose a practical NLI method but rather to make the theoretical claim that mastering the LM objective perfectly implies acquiring a full model of entailment. The inability of our distributional entailment test to distinguish entailment from near contradiction means the reconstruction of semantics it would extract from an idealized, perfect LM could still be fundamentally lossy.
Future work should investigate whether distributional semantics must fundamentally confuse entailment and near contradiction or whether there is some other way to distinguish them with form alone.

\subsection*{The Conceptual Problem}

The edge case that breaks \autoref{thm:informative} is simple to describe conceptually.
Our entailment test attempts to use the co-occurrence probability $p(xy)$ of two sentences $x, y$ to infer something about their semantic relationship.
Under a Gricean speaker, the probability of a redundant pair of utterances $x, y$ should be ${\sim}0$ and the probability of contradictory utterances $x, y$ should be exactly $0$.
The issue is when $y$ nearly contradicts $x$, e.g.:
\begin{align*}
x &= \texttt{I'm not in North America.} \\
y &= \texttt{I'm in a US state.}
\end{align*}
$y$ nearly contradicts $x$ because they are both satisfiable only in worlds where the speaker is in Hawaii, which we assume $p(w)$ makes unlikely. 
In such instances of near contradiction, it is possible that $p(xy)$ is slightly above 0 (like for entailment), and thus \autoref{thm:informative} detects entailment incorrectly.

\subsection*{Detailed Problem and Revised Analysis}

The technical problem in the original proof of \autoref{thm:informative} was that \autoref{lem:sum-entails} was applied with unmet preconditions.
Specifically, it assumed that the speaker utility $I_\ell(z; w)$ is at least $0$ for all utterances $z$ and worlds $w$, but, in fact, this utility is ${-}\infty$ in worlds where $z$ is false.
Near contradictions can then achieve an average exponentiated information content of $1$ because $\exp({-}\infty)$ in many worlds is balanced out by large positive information in a few worlds.
Formally, let $Y = \den{x} \cap \den{y}$ be the worlds where $x, y$ are both true.
We show in \autoref{app:erratum} that the original entailment test is 0 when
\begin{equation}
    p(Y) I(Y) = I , \label{eq:revised-cond}
\end{equation}
where $I(Y)$ measures the information content of $y$ given $x$ and $I$ is a constant reflecting $0$ information. We can see that \eqref{eq:revised-cond} has two distinct solutions:

\paragraph{Entailment Solution.} As expected, \eqref{eq:revised-cond} is satisfied when $x$ entails $y$: $p(Y) = 1$ and $I(Y) = I$.

\paragraph{Near-Contradiction Solution.} Assume for simplicity that $I_\ell(y \mid x; w) = c$ for all $w \in Y$.
If $y$ nearly contradicts $x$, $p(Y)$ is small because there are very few contexts where $x, y$ are both true. On the other hand, $I(Y)$ is large because $y$ is very informative when it is true. It is possible to calibrate $Y$ such that these factors multiply to $I$.

\begin{figure}[!t]
    \centering
    \includegraphics[width=\columnwidth]{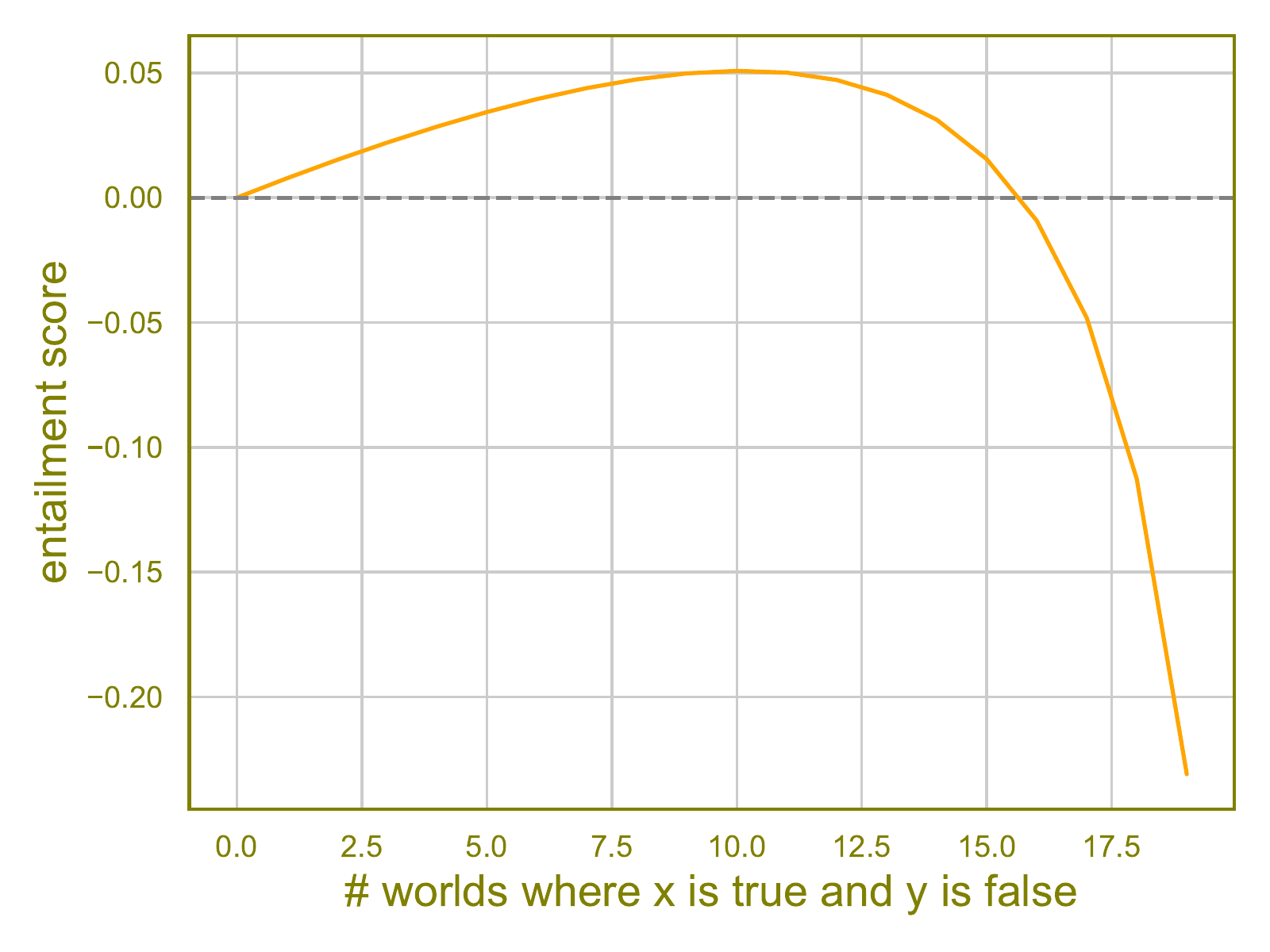}
    \captionsetup{labelfont={color=\erratumcolor}}
    \caption{\color{\erratumcolor} The entailment test score between $x$ and $y$ as a function of the number of worlds where $x$ is true but $y$ is false. Entailment (left intercept with $0$) and near contradiction (right intercept with $0$) look the same!}
    \label{fig:counterexample}
\end{figure}

\autoref{fig:counterexample} illustrates these two solutions using the Gricean speakers from \autoref{sec:learnability}.
For two utterances $x, y$, we vary the number of worlds where $x$ is true but $y$ is false, ranging from entailment to contradiction.
The test score crosses 0 twice: for entailment on the left and near contradiction on the right.

\color{black}
\pagebreak

\maketitle

\begin{abstract}
Language models are often trained on text alone, without additional grounding. There is debate as to how much of natural language semantics can be inferred from such a procedure. We prove that entailment judgments between sentences can be extracted from an ideal language model that has perfectly learned its target distribution, assuming the training sentences are generated by Gricean agents, i.e., agents who follow fundamental principles of communication from the linguistic theory of pragmatics. We also show entailment judgments can be decoded from the predictions of a language model trained on such Gricean data. Our results reveal a pathway for understanding the semantic information encoded in unlabeled linguistic data and a potential framework for extracting semantics from language models.
\end{abstract}

\section{Introduction}

Recent advances in building computational models of language have been powered by \emph{distributional semantics}: the idea that the possible surrounding contexts for a text span encode its meaning \citep{firth1957synopsis}.
% For example, word2vec vectors debatably encode semantic analogies between words after being trained to predict the immediate context of individual words \citep{mikolov2013efficient}.\tal{I would delete the analogy stuff, I don't find the empirical evidence that compelling (I guess I'm on the skeptical side of the ``debatably''.)}
In particular, large pretrained language models \citep[LMs;][]{peters-etal-2018-deep, devlin-etal-2019-bert, gpt3} have become an integral part of NLP systems: the representations that emerge from training to predict missing words in a text are empirically useful for natural language understanding tasks.
% \tal{This is the important part -- I would focus on modern LM instead of the Firth and analogies stuff.}

% But, are these models really learning to represent the meaning of text, and, if so, what part of their training data enables it?
% \citet{radford2019language} conjecture such an ability emerges from $\langle$task, input, output$\rangle$ tuples encountered in the training data. But such explicit supervision will be extremely sparse, suggesting the model might be leveraging other, more indirect, semantic signal \citep{zhang-hashimoto-2021-inductive}.

Despite this empirical progress, \citet{bender-koller-2020-climbing} argue LMs cannot learn to understand the semantics of sentences. This is because of a mismatch between the LM training objective---predicting missing words in text (``form'')---and \citeauthor{bender-koller-2020-climbing}'s conception of meaning as the relation of a sentence to the external world.
% In contrast, they conceptualize the meaning of a sentence as the \emph{truth conditions} the world must satisfy for the sentence to describe it.\footnote{Or more generally, as communicative intent.}
Thus \citeauthor{bender-koller-2020-climbing} claim ``that the language modeling task, because it only uses form as training data, cannot in principle lead to learning of meaning.''

In this paper, we argue meaning \emph{can} be learned from form because the communicative goals of human authors encode semantic information in unlabeled text. We show how this semantic information can be extracted to resolve semantic relations between sentences (e.g., whether one sentence entails another): in this inferentialist sense, ideal LMs encode the meaning of sentences.
% and that these artifacts are sufficient in principle to enable extracting meaning from text.
This argument has been raised speculatively by others \citep{blog2020, potts2020, foundation-models}, but we will rigorously justify it here with formal results.

To give the simplest (and least general) illustration of our argument, we first assume training data is generated by overly idealized \emph{uniformly truthful} speakers: agents who decide what to say by picking sentences they consider true uniformly at random.\footnote{Studying the ability of LMs to understand programming language semantics, \citet{merrill2021provable} make a similar assumption that programmers are more likely to write true assertion statements than false ones.}
This very coarsely captures human authors' goal of being informative (rather than misleading) to their listeners \citep{grice1975logic}.
In \autoref{thm:uniform}, we prove a sentence $x$ entails sentence $y$ if and only if, after uttering $x$, a uniformly truthful speaker is just as likely to say $y$ as to repeat $x$. Thus, entailment semantics can be extracted from probabilistic languages generated by uniformly truthful speakers.

Uniformly truthful speakers are not a realistic model of humans: while humans favor true sentences to false ones \citep{grice1975logic}, not all true sentences are equally likely to be produced.
It is a common principle in linguistic theories of pragmatics that
human speakers choose their utterances in order to balance two competing objectives: (a) conveying information to their listener and (b) brevity \citep{levinson1983pragmatics, grice1975logic}.
% However, \autoref{thm:cia-sta} shows that a generalized relation holds for a more general class of \emph{independently truthful} speakers who only produce true sentences and where adjacent sentences are conditionally independent given the world. Specifically, $x$ entails $y$ if and only if
% \begin{equation*}
%     \frac{p(xy)}{p(y)} = \frac{p(xx)}{p(x)} .
% \end{equation*}
% Yet conditional independence between sentences is also a strong simplifying assumption, since human speakers aim for each new sentence to convey new information \emph{given} what has already been said.
We define a class of \emph{Gricean} speakers who optimize for these objectives, 
and prove in \autoref{thm:informative} that $x$ entails $y$ if and only if a simple equation holds in terms of text probabilities produced by such speakers. Thus, entailment semantics can be decoded from probabilistic languages generated by Gricean speakers.

The previous results assume access to a language's ideal likelihood function, but, in practice, one only ever receives a corpus \emph{sampled} from the language.
Moving to the corpus setting, we analyze how much data allows approximately computing our derived entailment test using probabilities estimated from sentence frequencies in a corpus. We find that the corpus size needed to guarantee the entailment test holds approximately is inversely related to the likelihood of the sentences.
% Applying our theory,
We estimate that approximating the entailment test between $4$-word sentences using corpus frequencies is possible with $\sim10^{10}$ sentences, about the size of the GPT-3 training data \citep{gpt3}. On the other hand, approximating the entailment test for $10$-word sentences should be possible with $\sim10^{17}$ sentences, or $\sim 10^{7}$ GPT-3 corpora. Thus, extracting entailment judgments using corpus frequencies requires an infeasible amount of data---even by modern NLP standards.

To overcome this limitation, one might hope to use probabilities estimated by LMs to extract entailment judgments between longer sentences that are rare even in a large corpus. With synthetic data generated by Gricean speakers, we find that entailment can be decoded from n-gram LM predictions to some extent. However, we speculate that current neural LMs may not score the probability of rare text well enough to enable decoding entailment judgments between natural language sentences.

In summary, our main contribution is to show a correspondence between the semantics of text and its likelihood, assuming the likelihood function matches models of human text production from linguistic theory. Determining whether a sentence in a probabilistic language entails another sentence can be reduced to modeling the probabilities of strings in the language. In practice, entailment judgments between very short sentences can be extracted from corpus frequencies, but this becomes infeasible for slightly longer sentences.
LMs can in principle be used to extrapolate the likelihood of longer strings, but we hypothesize current LMs are not well-suited for doing so well enough to enable extracting entailment from natural language.
Our theory demonstrates a formal sense in which unlabeled text data encodes linguistic meaning and makes quantitative predictions for (a) how to extract semantics from text corpora and (b) how much data this requires.

\section{Definitions}

\subsection{Sentences and Worlds}

Let $\mathcal X$ be a finite set of sentences, and $\mathcal W$ a countable\footnote{Our results extend to uncountable sets of world states if entailment is relaxed to hold almost surely (cf. \autoref{sec:continuous}). Alternatively, our results apply as-is if we assume a countable set of equivalence classes over uncountably many worlds.} set of possible world states.
A sentence $x$ is a string whose denotation $\den{x}$ is a \emph{proposition}, i.e., a set of world states ($\subseteq \mathcal W$) where $x$ is true.
Following standard conventions in formal semantics \citep[cf.][]{Heim1998}, the set $\den{x}$ can be equivalently viewed as a function mapping a world state $w$ to $\{0, 1\}$ that indicates whether $x$ is true in $w$, which we will write as $\den{x}(w)$. We imagine $w$ to encode a partial description of the world, much like the concept of a \emph{situation} in formal semantics \citep{sep-situations-semantics}. For simplicity, we assume an individual's subjective belief state can be modeled as the unique, maximal $w$ that fully describes the facts which they believe to be true.

\paragraph{Example} $x = \texttt{John has at least two cats}$. Let $\mathcal W = \{ w_0, w_1, w_2, w_3 \}$ be the set of possible worlds, where $w_n$ denotes the state in which John has $n$ cats. Then $\den{x} = \{w_2, w_3\}$, because John has at least two cats in these worlds. Furthermore, it holds that $\den{x}(w_2) = 1$, but $\den{x}(w_1) = 0$.

\subsection{Speakers and Texts}

% Let $xy$ be the concatenation of two strings $x,y \in \mathcal X$.
We refer to a sequence of sentences $z \in \mathcal X^*$ as a \emph{text}.\footnote{Where $\mathcal X^*$ denotes the Kleene star closure of $\mathcal X$.} The meaning of a text is the set of worlds consistent with all its sentences, i.e.,
\begin{equation*}
    \den{z} = \bigcap_{t=1}^{\abs{z}} \den{z_t} .
\end{equation*}
We will imagine that a text $z \in \mathcal X^*$ is produced by iteratively sampling $z_t \in \mathcal X \cup \{ \omega \}$ from a \emph{speaker model} $p(z_t \mid z_{<t}, w)$.
% until the empty string is produced, i.e., for some $T$, $z_T = \omega$.
$p(z_t \mid z_{<t}, w)$ represents the probability of saying sentence $z_t$ with belief state $w$ after having said $z_1\cdots z_{t-1}$.
Let $\omega \not\in \mathcal X$ be a special \emph{end of sequence} token satisfying $\den{\omega} = \mathcal W$.
% Once $z_T = \omega$ for some $T$, we define $p(\epsilon \mid z\omega, w) = 1$ for any $z, w$, where $\epsilon$ represents the empty string.
We refer to any text ending with $\omega$ as \emph{complete}.
Given a world $w$, an incomplete text $z \in \mathcal X^*$ or complete text $z \in \mathcal X^* \omega$ has \emph{conditional probability}
\begin{equation*}
    p(z \mid w) = \prod_{t=1}^{\abs{z}} p(z_t \mid z_{<t}, w) .
\end{equation*}
The conditional probability of an incomplete text represents the probability of observing $z$ as the prefix of a text written by a human with beliefs $w$.
In contrast, the probability of a complete text represents the probability that a speaker produces $z$ and no further text.
The conditional distribution $p(z \mid w)$ cannot be observed directly by a LM, since $w$ is a latent variable missing from the training data. Rather, a LM has access to texts that have been generated by speakers across many possible belief states. Mathematically, this can be expressed by saying a LM's target distribution is a \emph{marginal} distribution over $z \in \mathcal X^* \cup \mathcal X^* \omega$ according to some \emph{prior} distribution over worlds $p(w)$:
\begin{align*}
    p(z)
    &= \EX_{w \sim p(w)} \left[ p(z \mid w) \right] \\
    &= \EX_{w \sim p(w)} \left[ \prod_{t=1}^\infty p(z_t \mid z_{<t}, w) \right] .
\end{align*}

% A sequence 
% If the each text has at most two sentences $x$ and $y$, then the \emph{target distribution} for a LM to learn is
% \begin{align*}
%     p(xy)
%         &= \sum_w p(xy, w)
%         % &= \sum_w p(xy \mid w) p(w)
%         = \sum_w p(x \mid w) p(y \mid x, w) p(w) .
% \end{align*}
% Informally, we are analyzing the ideal LM that has perfectly mastered its training objective: modeling the probability of texts.
The prior $p(w)$ represents the probability that a speaker contributing to the corpus will have belief state $w$---we make no assumptions about its form besides that $p(w) > 0$ for all $w \in \mathcal W$, and, for every sentence, there is some world state that makes that sentence true. In contrast to $p(z)$, which corresponds to the expected corpus frequency of $z$, we denote by $p(\den{z})$ the probability that $z$ is true.\footnote{The notation explicitly represents the probability mass assigned to the set of worlds where $z$ is true.}

\paragraph{Example}
Let $z$ be the $2$-sentence text:\footnote{Text taken from the \href{https://en.wikipedia.org/wiki/Kr\%C3\%A1kum\%C3\%A1l}{Wikipedia page} for the skaldic poem \emph{Krákumál}, written in Ragnar's voice.}
\begin{align*}
z_1 &= \texttt{We swung our swords.} \\
z_2 &= \texttt{That was ever so long ago.}
\end{align*}
% \begin{equation*}
%     z = \texttt{We swung our swords. That was ever so long ago...}
% \end{equation*}
Let $p$ be the distribution of all possible English web texts.
The marginal probability $p(z)$ can be decomposed across many possible worlds. One such world $w_1$ might be the world where the speaker is the semi-legendary Viking hero Ragnar Loðbrók (in modern English translation);  another world $w_2$ might be the perspective of a Reddit user reviewing a coffee maker. Each of these worlds corresponds to one term in a sum over all worlds. We expect $p(z \mid w_1)$ to be higher than $p(z \mid w_2)$ since it is more likely for a medieval literary character to utter $z$ than a modern product reviewer. Finally, $p(z \mid w_1)$ can be factored as
\begin{equation*}
    p(z_1 \mid w_1) p(z_2 \mid z_1, w_1) .
    % = p(\texttt{We swung...swords.} \mid w_1) p(\texttt{That...long ago...} \mid \texttt{We swung...swords.}, w_1) .
\end{equation*}
In contrast to $p(z)$, which counts all contexts where $z$ is the beginning of a longer text, $p(z \omega)$ measures the frequency of $z_1z_2$ followed by nothing else.

% We will also use the notation $p(y \mid x)$ to mean the probability of the string $y$ conditioned on the prefix $x$, i.e.,
% \begin{equation*}
%     p(y \mid x) = \frac{p(xy)}{\sum_{y'} p(xy')} .
% \end{equation*}

\subsection{Distributional and Semantic Relations} 
\paragraph{Distributional Relations} A \emph{distributional relation} $d$ is a relation over sentences $x$ and $y$ defined in terms of likelihood of different texts under some distribution $p$.
Let $d_p(x, y)$ be the value of the distributional relation $d$ between sentences $x, y$ according to distribution $p$.
If we train an LM on texts sampled from a target distribution $p$, the LM estimates a predictive distribution $\hat p$. Thus, any LM parameterizes $d_{\hat p}$: an instantiation of the distributional relation $d$ with respect to the probabilities learned by the LM.
If the LM perfectly approximates $p(x)$ for all $x$, then $d_{\hat p} = d_p$ by construction.

\paragraph{Example} Define the distributional relation $d$ (with respect to some distribution $p$) such that $d_p^>(x, y) \iff p(x) > p(y)$. $d^>_p(x,y)$ says $x$ is more likely than $y$ according to $p$. If $\hat p$ represents LM predictions trained on the target distribution $p$, than $d^>_{\hat p}(x, y)$ says whether the LM predicts a sentence $x$ is more likely than another sentence $y$.

\paragraph{Semantic Relations} In contrast, a semantic relation between $x$ and $y$ is a relation defined in terms of their denotations $\den{x}$ and $\den{y}$. We will focus on the key semantic relation of entailment:
\begin{definition}
For two sentences $x, y \in \mathcal X$, $x$ entails $y$ if and only if $\den{x} \subseteq \den{y}$.
\end{definition}
% \begin{definition}[Contradiction]
% For $\mathcal A, \mathcal B \subseteq \mathcal W$, we say $\mathcal A$ contradicts $\mathcal B$ if and only if $p(\den{xy}) = 0$.
% \end{definition}
It is not clear \emph{prima facie} if LMs can represent entailment relations. However, it could be that a semantic relation $s$ can somehow equivalently be written as a distributional relation $d_p$. If so, a LM that perfectly approximates $p$ could be understood to encode $s$, since $s$ can be extracted from $\hat p$ via $d_{\hat p}$.

Formally, we can ask if a semantic relation can be alternatively expressed as a distributional relation by analyzing if there exists an isomorphism between a semantic relation $s(\den{x}, \den{y})$ and some distributional relation $d_p(x, y)$:
\begin{definition}[Isomorphism] \label{def:isomorphism}
    A semantic relation $s$ is isomorphic to a distributional relation $d$ under speaker $p$ if and only if, for all $x, y \in \mathcal X$,
    \begin{equation*}
        s(\den{x}, \den{y}) \iff d_p(x, y) .
    \end{equation*}
\end{definition}
If \autoref{def:isomorphism} holds under a speaker model $p$, then predicting whether $s$ holds between two sentences is reducible to perfectly modeling the probabilities of texts generated by $p$.
Our goal going forward will be to derive distributional relations isomorphic to entailment assuming $p$ models the goals of humans when they produce text.

\section{Uniformly Truthful Speakers} \label{sec:uniform-true}

We start by illustrating our research question and technical approach assuming an overly simple model of humans as \emph{uniformly truthful} speakers.
A uniformly truthful speaker chooses a sentence to produce by selecting one of the true sentences that holds in their belief state uniformly at random. This very coarsely captures the property of natural language pragmatics that subjectively true sentences tend to be more likely than false ones, although it does not account for many other factors that influence human speech patterns in complex ways \citep{grice1975logic}.\footnote{LMs sometimes generate objectively false statements \citep{truthfulqa}, presumably due to the occurrence of such facts in their training data. This is actually consistent with a uniform truthfulness assumption, which only requires that speakers only produce sentences they believe are true, not sentences that are actually true in some objective sense.}
Let $n(w)$ be the number of \emph{sentences} true in world $w$. We can formally define a uniformly truthful speaker as follows:
\begin{definition}
A speaker $p$ is \emph{uniformly truthful} if, for all sentences $x \in \mathcal X \cup \{ \omega \}$,
\begin{equation*}
    p(x \mid w) = \frac{\den{x}(w)}{\sum_{x'} \den{x'}(w)} = \frac{\den{x}(w)}{n(w)} .
\end{equation*}
\end{definition}
In other words, $p$ uniformly spreads probability mass across all sentences that are true in world $w$.
% Under the uniformly truthful speaker, it is easy to see that contradiction is isomorphic to a distributional relation, because the speaker never produces a sentence that is false in, and any sentence that is never produced is false.
% \begin{remark}
% If $p$ is a uniformly truthful speaker, then contradiction is isomorphic to the distributional relation $d_p(x, y) \iff p(xy) = 0$.
% \end{remark}
We will show that, if the corpus consists of text written by uniformly truthful speakers, entailment can be decided by a distributional relation. The following lemma will be a core technical tool in our analysis. Informally, it is useful because it establishes a correspondence between relations over sets of worlds and probabilities.
\begin{lemma} \label{lem:sum-entails}
Let $\mathbbm{1}_{\mathcal S}$ be the indicator function for set $\mathcal S$.
For sets $\mathcal A, \mathcal B$ such that $\mathcal A \subseteq \mathcal B \subseteq \mathcal W$,  and $c : \mathcal W \to \mathbb{R}_+$, $\mathcal A = \mathcal B$ if and only if \begin{equation*}
    \sum_{w \in \mathcal W} \mathbbm{1}_{\mathcal A}(w) c(w) = \sum_{w \in \mathcal W} \mathbbm{1}_{\mathcal B}(w) c(w) .
\end{equation*}
\end{lemma}
\begin{proof}
We will prove that $\mathcal B \subseteq \mathcal A$ by contradiction. Assume there exists $w \in \mathcal B$ such that $w \not \in \mathcal A$. Then the right sum contains the positive term $c(w)$, while the left sum does not. Because all terms in the right sum are positive, the left sum must contain at least one term $c(w')$ that the right sum does not. Thus, $w' \in \mathcal A$ but $w' \not\in \mathcal B$. But this has violated our assumption that $\mathcal A \subseteq \mathcal B$.
\end{proof}

We now use \autoref{lem:sum-entails} to derive a simple distributional relation that is isomorphic to entailment.
\begin{theorem} \label{thm:uniform}
If $p$ is a uniformly truthful speaker, then entailment is isomorphic to a distributional relation. Specifically, for all sentences $x,y \in \mathcal X$,
\begin{equation*}
    \den{x} \subseteq \den{y} \iff p(xy) = p(xx) .    
\end{equation*}
\end{theorem}

\begin{proof}
$d_p(x, y)$ holds if and only if
\begin{align*}
    p(xy) &= p(xx) \\
    \EX_w \left[  \frac{\den{x}(w) \den{y}(w)}{n(w)^2} \right] &= \EX_w \left[  \frac{\den{x}(w) \den{x}(w)}{n(w)^2} \right] \\
    \EX_w \left[  \frac{\den{x}(w) \den{y}(w)}{n(w)^2} \right] &= \EX_w \left[  \frac{\den{x}(w)}{n(w)^2} \right] .
\end{align*}
An expectation in a countable space is a sum weighted by probability masses.
So, by \autoref{lem:sum-entails}, this holds iff $\den{x} = \den{xy} = \den{x} \cap \den{y}$.
We conclude $p(xy) = p(xx)$ if and only if $\den{x} \subseteq \den{y}$.
\end{proof}

A similar proof suffices to show that the following isomorphism also holds:

\begin{corollary} \label{cor:uniform-sim-informative}
If $p$ is a uniformly truthful speaker, the following isomorphism holds for all $x,y \in \mathcal X$:
\begin{equation*}
    \den{x} \subseteq \den{y} \iff p(xy) = p(x\omega) .
\end{equation*}
\end{corollary}

\subsection{Discussion}
Uniformly truthful speakers resemble humans in that they mimic the \emph{tendency} of humans to tell the truth about what they believe. However, they are clearly too simple to account for human speech patterns. Most crucially, humans generally aim to produce informative speech, rather than sampling true sentences at random. More fundamentally, natural language has a countably infinite number of possible sentences, so a uniform distribution over all true sentences is not even mathematically well-defined. These limitations motivate our more involved analysis of Gricean speakers, which will adapt the technical tools used in this section.

\section{Gricean Speakers} \label{sec:informative}

In this section, we will define a new class of speakers who pick sentences in order to be informative to their listener, while also trying to be concise.
To do this, we will draw on information theory to formalize what it means for a speaker to be informative.
We will then derive a distributional relation that is isomorphic to entailment for Gricean speakers, which is a generalization of the relation for uniformly truthful speakers from \autoref{sec:uniform-true}.

\subsection{Definition}

\paragraph{Information} The first step towards formalizing Gricean speakers is to define a notion of the semantic information contained in a sentence.
% Let $\mathcal X^*$ be the Kleene star of $\mathcal X$, i.e., the set of sequences of sentences.
We formalize a listener $\ell(w \mid z)$ as the inverse of a speaker: Given a text $z \in \mathcal X^*$, a listener produces a distribution over possible world states. Then, in a given world $w$ we can define the information that a text conveys to the listener as the reduction in the number of bits needed to transmit $w$ to $\ell$ after they have read $z$ compared to before they have read $z$.
% \will{Should rework this notation $I_\ell(w \mid z)$ and cut conditional info?? Maybe not worth}
\begin{definition} \label{def:info}
The information content of a text $z \in \mathcal X^* \cup \mathcal X^*\omega$ to a listener $\ell(w \mid z)$ is\footnote{For convenience, we let $\log 0 = -\infty$ and $\infty - \infty = 0$.}
\begin{align*}
    I_\ell(z; w) = \log \ell(w \mid z) - \log \ell(w) .
\end{align*}
\end{definition}
In other words, the information content of a text is the reduction in $\ell$'s code length for the world after having read the text compared to beforehand.
We can naturally extend \autoref{def:info} to measure the \emph{conditional information} conveyed by sentence $y$ given that $x$ has already been produced:
\begin{definition} \label{def:cond-inf}
The information content of $y \in \mathcal X^* \cup \mathcal X^*\omega$ given $x \in \mathcal X^*$ to a listener $\ell(w \mid z)$ is
\begin{align*}
    I_\ell(y \mid x; w)
    &= I_\ell(xy; w) - I_\ell(x; w) \\
    &= \log \ell(w \mid xy) - \log \ell(w \mid x) .
\end{align*}
\end{definition}
% \noindent It follows from \autoref{def:cond-inf} that
% \begin{equation*}
%     I_\ell(y \mid x; w) = \log \ell(w \mid xy) - \log \ell(w \mid x) .
% \end{equation*}

\paragraph{Informative Speaker} We now define a Gricean speaker in terms of $I_\ell$. Our definition generalizes the rational speech acts model \citep{goodman2016pragmatic}, but makes weaker assumptions about the listener and allows a dynamic semantics where later sentences can condition on previous ones \citep{lewis1979scorekeeping,kamp1981theory,heim1982semantics}.
% \footnote{In \autoref{sec:rsa}, we further discuss the connection to RSA.
% % We also analyze a related but less realistic model of speakers who are non-dynamically informative in \autoref{sec:ind-truthful}.
% }
We define an utterance's utility as a convex combination of its information content and its cost to produce, operationalizing the Gricean idea that speakers pick utterances by weighing their informativeness against their cost. The cost function $c : \mathcal X^* \cup \mathcal X^*\omega \to \mathbb R$ can be any measure of sentence complexity (e.g., length) satisfying $c(xy) = c(x) + c(y)$ for $x, y \in \mathcal X^* \cup \mathcal X^*\omega$.\footnote{This is satisfied when $c(x)$ is the length of $x$, but also for other options like the corpus frequency of $x$ \citep{goodman2016pragmatic} or the depth of the syntactic tree of $x$.}
\begin{definition} \label{def:cont-informative}
A speaker $p$ is \emph{Gricean} if there exists a listener $\ell(w \mid z)$,
some $\alpha > 0$,
% an invertible function $f: \mathbb{R} \to \mathbb{R}$,
and a cost function $c$
% $c: \mathcal X^* \cup X^*\omega \to \mathbb{R}$
such that, for all $z \in \mathcal X^* \cup \mathcal X^* \omega$:\footnote{\color{\erratumcolor} To clarify, we assume $p(z \mid w)$ is locally normalized at each sentence rather than globally. However, our results also go through with global normalization as well.}
\begin{equation*}
    p(z \mid w) \propto \exp \left( \alpha I_\ell(z; w) - c(z) \right) .
    % p(y \mid x, w) \propto \exp \left( f(I_\ell(y \mid x; w)) - c(y) \right) .
\end{equation*}
Further, $\ell$ must satisfy the following for all $x \in \mathcal X^*$, $y \in \mathcal X \cup \{\omega\}$, and $w \in \mathcal W$,
\begin{equation*}
    I_\ell(y \mid x; w) = 0 \iff \den{x}(w) \to \den{y}(w) .
\end{equation*}
\end{definition}
% \will{$I_\ell(w \mid xy) = I_\ell(w \mid x)$, i.e., $\ell(w \mid xy) = \ell(w \mid x)$.}
In other words, the speaker must be trying to convey information about the state of the world to some listener who fully absorbs the semantic information in all sentences they have already heard: clarifying already established information will not benefit the listener. We can formalize this by deriving $p(y \mid x, w)$ for $x \in \mathcal X^*$ and $y \in \mathcal X \cup \{\omega\}$:
\begin{align*}
    p(y \mid x, w)
    &= \frac{p(xy \mid w)}{p(x \mid w)} \\
    % &= \exp \left( \alpha I_\ell(xy; w) - c(xy) - \alpha I_\ell(x; w) + c(x) \right) \\
    &\propto \exp \left( \alpha I_\ell(y \mid x; w) - c(y) \right) .
\end{align*}
Notably, the probability of $y$ given $x$ depends on the \emph{conditional information} of $y$ given $x$, which means only information conveyed by $y$ that is nonredundant with $x$ will make $y$ more likely.\footnote{From a technical perspective, the $\exp$ in \autoref{def:cont-informative} is justified by the fact that probabilities decompose multiplicatively, i.e., $p(xy \mid w) = p(x \mid w) p(y \mid x, w)$, but the information content and cost of text should decompose additively across different sentences. Applying basic exponent rules shows that \autoref{def:cont-informative} satisfies this desideratum.}

\subsection{Results}

Proofs are in \autoref{app:informative}.
Under a Gricean speaker, the cost of an utterance can be expressed:
\begin{restatable}{lemma}{lemCost} \label{lem:cost}
For any Gricean speaker $p$ and $x \in \mathcal X$,
\begin{equation*}
    \frac{p(x\omega)}{p(xx)} = \frac{\exp(c(x))}{\exp(c(\omega))} .
\end{equation*}
\end{restatable}
% Before proceeding to the main result, we note an interesting consequence of \autoref{lem:cost} for distributionally estimating the cost that speakers assign to utterances. The following is derived by taking the $\log$ of both sides in \autoref{lem:cost}:
\begin{corollarylem} \label{cor:rsa-cost}
Under a Gricean speaker, for all $x \in \mathcal X$,
$c(x) = \log p(x\omega) - \log p(xx) + c(\omega)$.
% \begin{equation*} 
%     c(x) = \log p(x\omega) - \log p(xx) .
% \end{equation*}
\end{corollarylem}
\autoref{cor:rsa-cost} says that a sentence is costly to the extent that it is unlikely to be repeated twice, giving an intuitive characterization of this quantity in terms of text probabilities. Now, we will use this characterization of cost to derive a distributional relation that is isomorphic to entailment.

\begin{restatable}{theorem}{thmInformative} \label{thm:informative}
Under any Gricean speaker $p$, entailment is isomorphic to a  distributional relation. Specifically, for all sentences $x,y \in \mathcal X$,
\begin{equation*}
    \den{x} \subseteq \den{y} \iff \frac{p(xy)}{p(x \omega)} = \frac{p(yy)}{p(y\omega)} .
\end{equation*}
\end{restatable}
If we allow our decision rule to depend on the cost function $c$ in addition to probabilities, we can simplify \autoref{thm:informative} as follows:

\begin{corollary} \label{cor:informative-no-cost}
Under any Gricean speaker $p$, for all sentences $x,y \in \mathcal X$, $\den{x} \subseteq \den{y}$ if and only if
\begin{equation*}
    \log p(x \omega) - \log p(xy) = c(y) - c(\omega) .
\end{equation*}
\end{corollary}

If we imagine $c(y) - c(\omega) = 0$ for a uniformly truthful speaker, we see the equation in \autoref{thm:informative} is a generalization of the equation in \autoref{thm:uniform}.

\subsection{Discussion}

Gricean speakers are a general enough model of humans speakers to capture the basic pragmatic principles influencing speech production. Thus, it is notable that \autoref{thm:informative} establishes a closed-form distributional relation isomorphic to entailment.

One conceptual limitation of Gricean speakers is that their simulated listener must fully consume information, such that redundantly conveying the same information twice will not lead to any information gain the second time.
This contrasts with real speech, where potential interpretation errors by the listener incentivize the speaker to be somewhat redundant \citep{degen2019redundancy}.
Mathematically, this would violate the axiom of \autoref{def:cont-informative} that
\begin{equation*}
    I_{\ell}(y \mid x; w) = 0 \iff \den{x}(w) \to \den{y}(w) .
\end{equation*}
Extending \autoref{thm:informative} to speakers who use redundancy to account for noise and interpretation errors is an interesting direction for future work.

Another interesting extension would be formalizing speakers who aim to be informative regarding some question under discussion, rather than being generally informative about $w$ \citep[cf.][]{goodman2015probabilistic}.
This could encompass both ``what'' questions that aim to clarify some aspect of the world, and ``why'' questions that aim to convey explanations for established facts.
% While intuitively this seems like it may complicate the mapping between form and meaning, we show in \will{Add appendix} that the isomorphism established in \autoref{thm:informative} still holds in this more general case, as long as the question under discussion is a ``what question'', i.e., a question that inquires about some part of $w$. On the other hand, our model is not general enough to formalize ``why questions'' that aim to explain the validity of some claim, rather than convey new information about the world.

\section{Decoding Entailment from Empirical Text Frequencies} \label{sec:finite-analysis}

We have so far shown that entailment judgments can be extracted from the sentence probabilities in the ideal distribution $p(z)$. What happens if, more practically, we estimate the probability of a sentence by its frequency in a large corpus sampled from $p(z)$? We prove this method enables feasible extraction of entailment judgments between very short sentences, but the corpus size may become intractably large for longer sentences.

Imagine we have a finite corpus of \emph{iid} sentences $\{Z_i\}_{i=1}^n$, each sampled from $p(z)$.
Let $\hat p(z)$ be the empirical frequency of a text $z$ in the corpus, i.e.,
if $\pi(z, z')$ returns whether text $z$ is a prefix of text $z'$,
\begin{equation*}
    \hat p(z) = \frac{1}{n} \sum_{i=1}^n \pi(z, Z_i) .
\end{equation*}

Since $p(z)$ encodes entailment via our extraction rules, $\hat p(z)$ will encode entailment between sentences if $\hat p(z)$ is close to $p(z)$. A naive notion of closeness is to guarantee, for all $\epsilon$, there exists some number of texts $n$ such that, with high probability, $\abs{ p(z) - \hat p(z) } < \epsilon$.
But this notion is not strict enough: if $p(z)$ is small, this difference will also be small, even if $\hat p(z)$ is not a good approximation of $p(z)$ on a relative scale. Instead, we want to guarantee that $\hat p(z) / p(z)$ converges to $1$, or, equivalently, that their difference as log probabilities converges to $0$.
This ensures that convergence will still be meaningful for low-probability sentences, which most sentences are in natural language.

Under this standard, rarer sentences take more samples to approximate. Define the sentence complexity $\mathfrak K_p(z) = \frac{1}{p(z)}$. We bound the approximation error in terms of $\mathfrak K_p(z)$.\footnote{Omitted proofs from \autoref{sec:finite-analysis} are in \autoref{sec:proofs}.}

\begin{restatable}{lemma}{lemLogConcentration} \label{lem:log-concentration}
For $z \in \mathcal X^* \cup \mathcal X^*\omega$ and $\delta > 0$, it holds with probability at least $1 - \delta - (1 - p(z))^n$ that
\begin{equation*}
    \abs{\log p(z) - \log \hat p(z)} \leq \sqrt{ \frac{\mathfrak K_p(z)}{\delta n} } .
\end{equation*}
\end{restatable}

To make this bound non-vacuous, $n$ must be large enough to counteract $\mathfrak K_p(z)$ and bring $(1-p(z))^n$ close to $0$.
Thus, good approximation requires fewer samples for more common sentences.
To get a more concrete view of the number of samples required to extract entailment judgments from an LM, we analyze $\mathfrak K_p(z)$ for Gricean speakers.\footnote{\autoref{sec:proofs} also analyzes uniformly truthful speakers.}

% Three conditions enable entailment from text frequencies from $n$ sentences from a uniformly truthful speaker:
% \begin{enumerate}
%     \item The text $xy$ must be frequent enough to ensure the bound holds with high probability.
%     \item The number of sentences $\abs{\mathcal X}$ must not be too large, or else the bound will become vacuous.
%     \item The truth probability $p(\den{xy})$ must not be too small, or else the bound will become vacuous.
% \end{enumerate}
% These criteria are achievable in toy languages with a small set of sentences. In natural language, though, most sentences are rare, so $n$ may have to be very large to enable decoding entailment from uniformly truthful texts.

% To go beyond analyzing a uniformly truthful speaker, we will need a more general characterization of $\mathfrak K_p(z)$.
Recall that we write $c(z)$ for the cost that a Gricean speaker assigns to producing a sentence $z$. For Gricean speakers, $\mathfrak K_p(z)$ is related to $c(z)$ as well as the probability $z$ is true.
\begin{restatable}{theorem}{thmLogBound} \label{thm:log-bound}
Assume that $p(z \mid w)$ is a Gricean speaker with respect to listener $\ell$ and $\den{z}(w) = 1 \iff I_\ell(z ; w) \geq 0$.
Let $g_p(x, y) = \log \frac{p(xy)}{p(x\omega)} - \log \frac{p(yy)}{p(y\omega)}$ . Let $q = 1 - \min \{ p(xy), p(yy) \}$. Then, for all $x, y \in \mathcal X$ such that $\den{xy}(p) > 0$, for all $\delta > 0$,
it holds with probability at least $1 - \delta - 4q^n$ that $\abs{g_p(x, y) - g_{\hat p}(x, y)}$ is at most
\begin{equation*}
    8\sqrt{ \frac{\exp( \max \{ c(xy), c(yy) \} )}{p(\den{xy})} \cdot \frac{1}{\delta n}} .
\end{equation*}
\end{restatable}

\autoref{thm:log-bound} says we can use text frequencies to decode entailment between sentences $x,y$ from a Gricean corpus, but the number of training sentences to guarantee this grows exponentially with the cost of $x$ and $y$. Thus, we probably cannot expect to extract entailment judgments from text frequencies except between \emph{very short} sentences.

\begin{figure}
    \centering
    \includegraphics[width=\columnwidth]{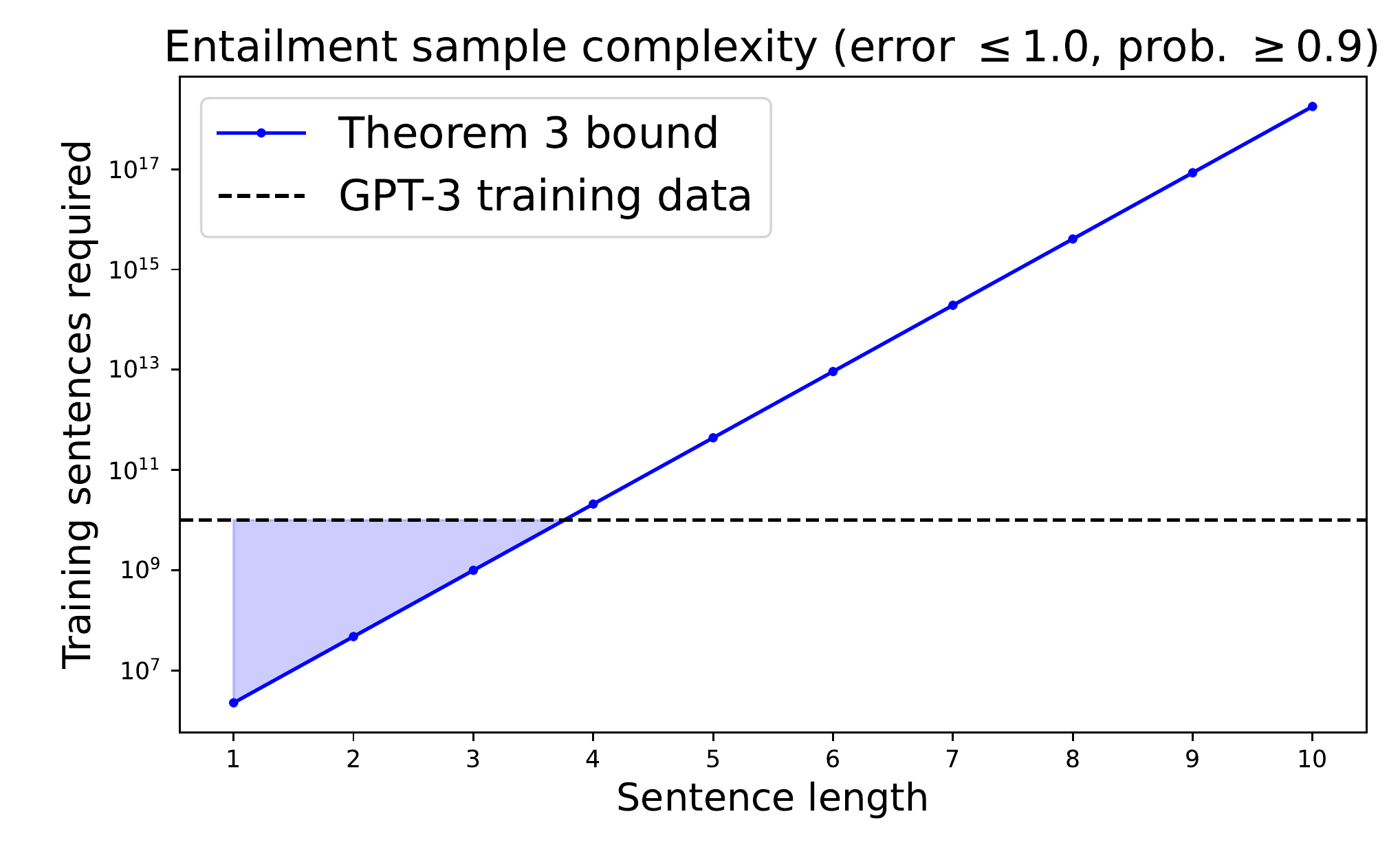}
    \caption{Estimated number of training sentences for guaranteeing $g_{\hat p}$ closely approximates $g_p$, where $\hat p$ is estimated using empirical text frequencies.}
    \label{fig:sample-complexity}
\end{figure}

We make this more quantitative in \autoref{fig:sample-complexity}, where we estimate the number of training sentences needed to ensure $g_p$ and $g_{\hat p}$ are close on sentences of length $\leq k$ as a function of $k$.
The main assumption behind this calculation is that a sentence's probability vanishes exponentially in its length, where the exponential base is the perplexity of the language. \autoref{app:estimation} documents the underlying assumptions in more detail.
\autoref{fig:sample-complexity} predicts $g_p$ and $g_{\hat p}$ can be made close for length-$4$ sentences using $\sim 10^{10}$ training sentences: about as much data as GPT-3 was trained on. In contrast, handling (still short) sentences of length $10$ can be done with $\sim 10^{17}$ training sentences, or $\sim 10^{7}$ GPT-3 corpora. Thus, relying solely on corpus frequencies is likely not a feasible way to extract entailment relations from text generated by Gricean speakers.

% An interesting direction for future work would be to analyze whether LMs with a strong model of compositional structure enable more sample-efficient entailment decoding than the sentence-level n-gram models analyzed here.

\section{Decoding Entailment from LMs} \label{sec:learnability}

We have just analyzed how many samples are necessary to decode entailment relations from the text frequencies in a finite corpus.
As shown by \autoref{thm:log-bound}, this approach will require intractably many samples for sentences of nontrivial length because longer strings will appear infrequently (if at all) in the corpus.
In order to estimate the probability of rare, longer, strings what if we use an LM to estimate $\hat p(z)$ instead of text frequencies?
Perhaps a smoothed LM should allow us to extrapolate $\hat p(z)$ well enough for long sentences to extract entailment judgments between them.
In this section, we briefly discuss some limitations of this approach.

It is tempting to take low LM perplexity as evidence that an LM estimates sentence probabilities well enough to approximately satisfy the isomorphism in \autoref{thm:informative}. After all, low test perplexity implies that $\hat p(z)$ is, on average, a good approximation of $p(z)$: if the perplexity is bounded below $\epsilon$, then the KL divergence $\mathrm{KL}(p, \hat p)$ is bounded below $\log \epsilon$.
% \begin{equation*}
%     \mathrm{KL}(p, \hat p) = \EX_{z \sim p(z)} \left[ \log p(z) - \log \hat p(z) \right] < \log \epsilon .
% \end{equation*}
$\epsilon$ decreases with the amount of training data $n$ at a rate between $\Omega(1/\sqrt{n})$ and $\Omega(1/n)$ \citep{wang2013consistency, li2021towards}.
Thus, with enough data, $\hat p(z)$ will closely approximate $p(z)$ for an average sentence $z$ in the training distribution.

But low error on an average $z$ does not establish entailment can be decoded from $\hat p$ because $d_{\hat p}$, as derived in \autoref{thm:informative}, depends on the text $z=yy$, which is very unlikely in natural language.\footnote{$yy$ is unlikely to be produced by a Gricean speaker because the second $y$ conveys no information.} Poorly estimating $p(yy)$ has little impact on $\mathrm{KL}(p, \hat p)$, so LMs trained to minimize $\mathrm{KL}(p, \hat p)$ have no reason to estimate $p(yy)$ well unless they are imbued with strong inductive biases. Thus, we expect that
% ---despite our theoretical results for their target distribution---
LMs trained with a standard cross-entropy loss may not produce reliable entailment judgments because they poorly estimate the probability of key valid (but unlikely) texts.\footnote{Future work should more carefully analyze how much data is required to extract complex entailment relations from LM predictions (rather than corpus frequencies). This is beyond the scope of the current project.} However, we find in the next section that they do succeed in the easier setting of small artificial languages and fully Gricean speakers.

% \section{Experiments}

% \subsection{Semantic Shift}
% Another conceptual issue with learning entailment from language modeling is the assumption that a static $p(z)$ exists in the first place \citep{bender-koller-2020-climbing}. Language changes over time, so by the time a LM has been trained to approximate $p$, $p$ may have shifted to some new distribution $p'$. This does not seem to 

\section{Experiments: Extracting Semantics from Simulated Gricean Corpora}

We test empirically whether we can extract entailment judgments from LMs trained on unlabelled text.\footnote{\url{https://github.com/viking-sudo-rm/formal-language-understanding}} Natural language corpora are unlikely to adhere exactly to our idealized assumptions about the speakers generating texts, so we generate the training corpora from a simulated Gricean speaker (see \autoref{sec:informative}). To make learning semantics more tractable with limited computation, we set $\abs{\mathcal W} = 3$ and restrict the vocabulary $\mathcal X$ to 7 utterances, each denoting one of the 7 non-empty subsets of $\mathcal W$. Each sentence in the training corpus is generated by sampling utterances from a Gricean speaker, conditioned on a uniformly sampled world state and the previously generated utterance, until the tautological utterance is generated. The semantic value of a sentence is taken to be the conjunction over all of its utterances. We set the rationality parameter $\alpha$ and the cost function heuristically (details in \autoref{sec:details}).

We generate training sets varying in size from 2 texts to 10M texts, and train two types of models on each: a simple empirical text frequency as described in Section \ref{sec:finite-analysis}, and a trigram model implemented using NLTK \citep{bird-2006-nltk}. Then for all sentence pairs $(x, y)$, where $x$ and $y$ have 6 utterances or fewer and each denotes a non-empty proposition, we compute $g_{\hat{p}}(x, y)$ from \autoref{sec:finite-analysis}. \autoref{thm:informative} shows that, under the true distribution $p$, $g_{\hat{p}}(x, y) = 0$ if and only if $x$ entails $y$.

\begin{figure}
    \centering
    \begin{subfigure}[t]{\columnwidth}
        \centering
        \includegraphics[width=\columnwidth]{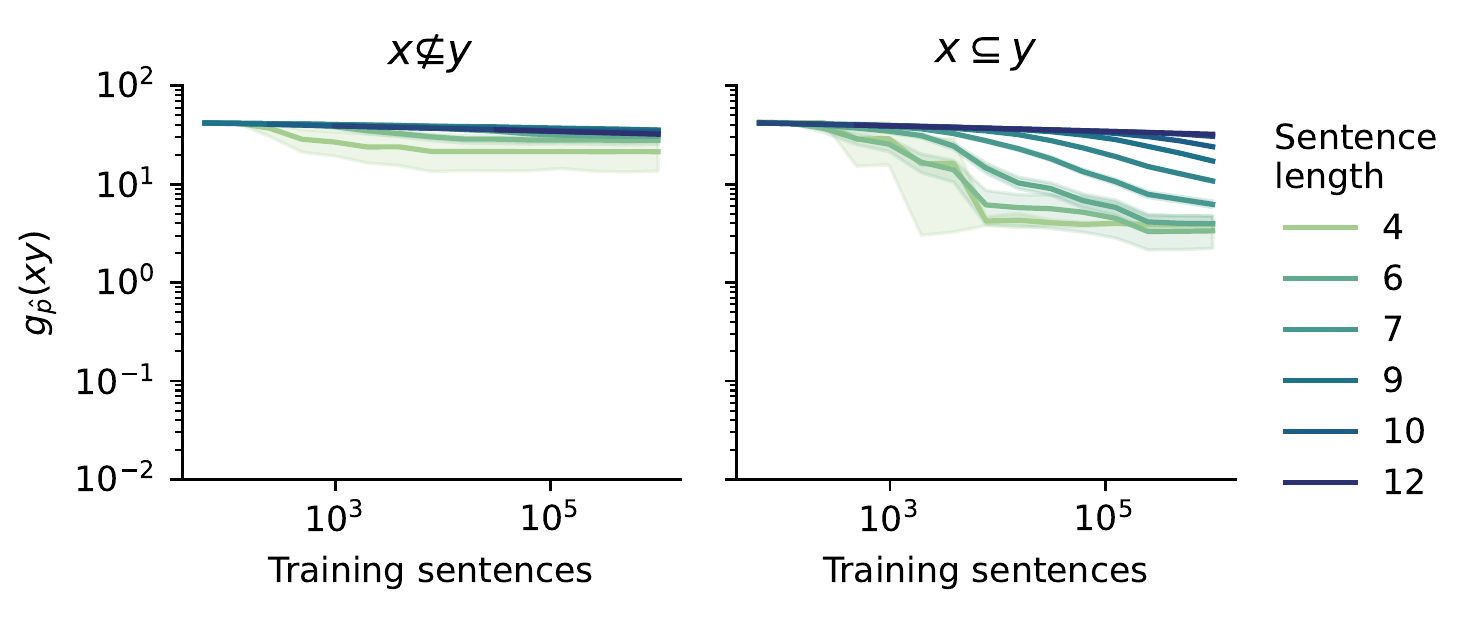}
        \caption{$\hat{p}(x)$ given by text frequency model.}
        \label{fig:exp_text_frequency}
    \end{subfigure}
    
    \begin{subfigure}[b]{\columnwidth}
        \centering
        \includegraphics[width=\columnwidth]{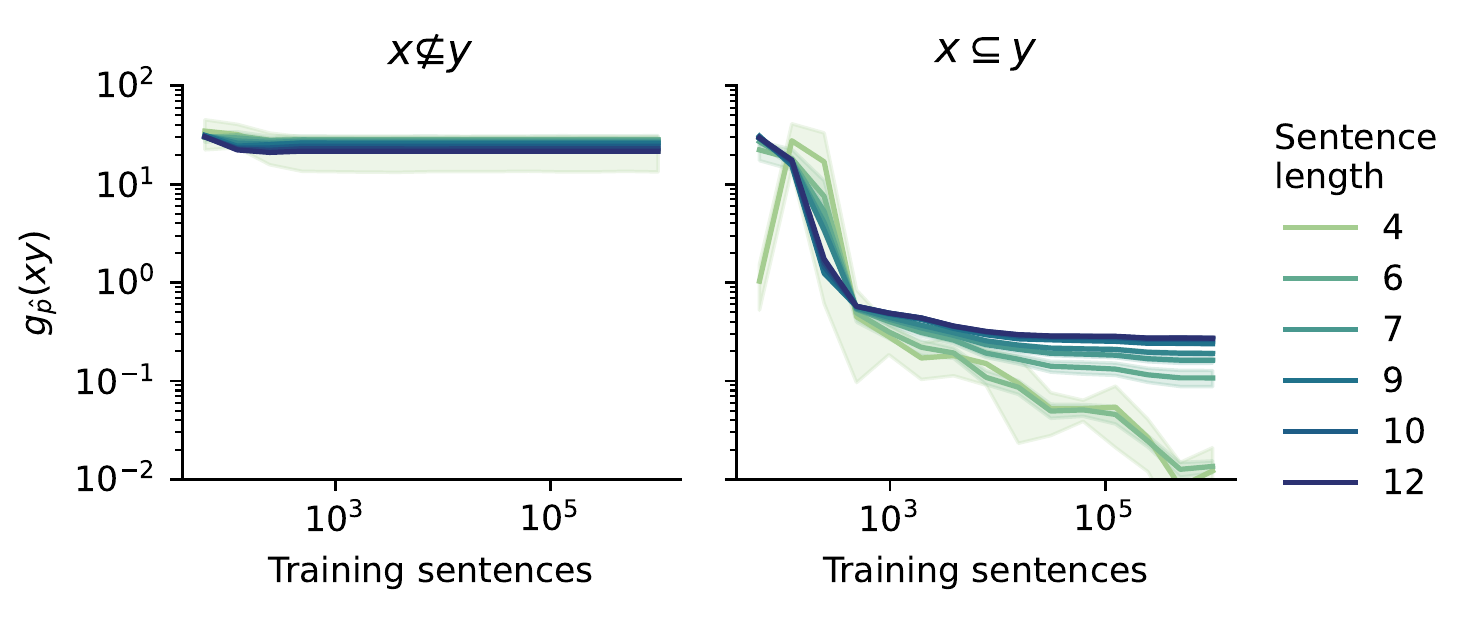}
        \caption{$\hat{p}(x)$ given by trigram model.}
        \label{fig:exp_ngram}
    \end{subfigure}
    \caption{Plot of $g_{\hat{p}}(x, y) = \log\frac{\hat{p}(xy)}{\hat{p}(x\$)} - \log\frac{\hat{p}(yy)}{\hat{p}(y\$)}$ as a function of the number of sentences in the training corpus and the length $\abs{xy}$. Given the true distribution $p$, $g_{p}(x, y) = 0$ iff $x$ entails $y$. We exclude pairs $x,y$ where both $xy$ and $yy$ are absent from the training data.}
    \label{fig:exp}
\end{figure}

The results are plotted in Figure \ref{fig:exp}. We arrive at the following conclusions:

\paragraph{Entailment relations can be extracted with greater-than-chance performance from LM predictions.} The value of $g_{\hat{p}}(x, y)$ is much closer to 0 on average for entailed pairs than for non-entailed pairs. This is predicted by Theorem 2.

\paragraph{The size of the corpus needed to extract entailment grows predictably with sentence length.} For entailed pairs, the average value of $g_{\hat{p}}(x, y)$ for shorter sentences approaches 0 more quickly with a large training corpus. This is in line with the predictions of Theorem 4.

\paragraph{Model inductive bias impacts the ease of extracting entailment.} Entailed and non-entailed pairs are better distinguished by the trigram model than the text frequency model. Specifically, $g_{\hat{p}}(x, y)$ is closer to 0 for the trigram model for a given amount of data, and the trigram model's predictions are less sensitive to sentence length.

\section{Generality of Extracting Semantics} \label{sec:generality}

Our main result that entailment judgments can be extracted from an ideal LM assumes the corpus was produced by Gricean speakers. While pragmatic theory supports this assumption, real human speakers are undoubtedly more complex. What if we relax the assumption that speakers are Gricean?
In \autoref{thm:existence} in \autoref{sec:existence}, we show that any semantic relation is isomorphic to some distributional relation as long as, for any pair of possible semantics, there is some text whose probability distinguishes between the two candidate semantics.

We take it to be uncontroversial that semantics influences speech production, so we interpret \autoref{thm:existence} to say all semantic relations are fully encoded in ideal LMs. In contrast to \autoref{thm:informative}, however, this result is nonconstructive, so we do not know which algorithm to use to decide entailment between two sentences, even though one exists.
Further, without further assumptions about the speaker, we cannot guarantee the extraction relation is efficiently computable or even computable at all.

% In CITE, we construct an adversarial case following \citet{merrill2021provable} where 

\section{Conclusion}

Given a general, linguistically motivated model of human text production, we proved that entailment judgments can be decoded from the likelihood function for texts because of semantic artifacts created by human authors.
We also showed empirically that entailment could be extracted n-gram LMs trained on simple formal languages.
Thus, we have given one explanation for why distributional information encodes semantic information \citep{firth1957synopsis} and how semantic relations are, in principle, extractable from LMs.
% The meaning of a sentence is often conceptualized as truth conditions that speakers represent cognitively, rather than as information contained within the text itself. 
% Thus, \citet{bender-koller-2020-climbing} argued LMs cannot learn semantics, since they are only trained on text, rather than the underlying truth conditions.
% Our results suggest the disconnect between form and meaning is not insurmountable:  In a large corpus of texts, we can even expect to extract this entailment information for short sentences. This suggests a theoretical pathway for the distributional hypothesis \citep{firth1957synopsis} that word cooccurences encode semantics: such a correspondence can arise as an artifact of the pragmatic goals of Gricean speakers.
It is an open question whether entailment judgments might be extractable from current large LMs, but we hypothesize that the complexity of natural language makes this substantially more challenging than with our synthetic experiments, and that the loss function and inductive biases of current neural LMs are not well suited for doing so without an infeasible amount of data.

A natural next step for future work is to test this hypothesis empirically by measuring whether entailment judgments can be extracted from large LMs using our theory. Similarly, it would be interesting to think about how LMs could be modified so that they can better pick up on the semantic information encoded in their training distribution.

\section{Acknowledgments}
This project benefited from informal discussions with
Chris Barker, Sam Bowman, Lucius Bynum, Kyunghun Cho, Tiwa Eisape, Yanai Elazar, Najoung Kim, Alexander Koller,
Vishakh Padmakumar, Chris Potts, Naomi Saphra, Sebastian Schuster, and Noah A. Smith.
We also thank the members of the CAP Lab and Semantics Group at NYU for their feedback.
This work was supported in part through the NYU IT High Performance Computing resources, services, and staff expertise.
It was funded by NSF award 1922658, and WM was supported by an NSF graduate research fellowship.

% Entries for the entire Anthology, followed by custom entries
\bibliography{anthology,references}

\begin{thebibliography}{23}
\expandafter\ifx\csname natexlab\endcsname\relax\def\natexlab#1{#1}\fi

\bibitem[{Bender and Koller(2020)}]{bender-koller-2020-climbing}
Emily~M. Bender and Alexander Koller. 2020.
\newblock \href {https://doi.org/10.18653/v1/2020.acl-main.463} {Climbing
  towards {NLU}: {On} meaning, form, and understanding in the age of data}.
\newblock In \emph{Proceedings of the 58th Annual Meeting of the Association
  for Computational Linguistics}, pages 5185--5198, Online. Association for
  Computational Linguistics.

\bibitem[{Bird(2006)}]{bird-2006-nltk}
Steven Bird. 2006.
\newblock \href {https://doi.org/10.3115/1225403.1225421} {{NLTK}: The
  {N}atural {L}anguage {T}oolkit}.
\newblock In \emph{Proceedings of the {COLING}/{ACL} 2006 Interactive
  Presentation Sessions}, pages 69--72, Sydney, Australia. Association for
  Computational Linguistics.

\bibitem[{Bommasani et~al.(2021)Bommasani, Hudson, Adeli, Altman, Arora, von
  Arx, Bernstein, Bohg, Bosselut, Brunskill, Brynjolfsson, Buch, Card,
  Castellon, Chatterji, Chen, Creel, Davis, Demszky, Donahue, Doumbouya,
  Durmus, Ermon, Etchemendy, Ethayarajh, Fei-Fei, Finn, Gale, Gillespie, Goel,
  Goodman, Grossman, Guha, Hashimoto, Henderson, Hewitt, Ho, Hong, Hsu, Huang,
  Icard, Jain, Jurafsky, Kalluri, Karamcheti, Keeling, Khani, Khattab, Koh,
  Krass, Krishna, Kuditipudi, Kumar, Ladhak, Lee, Lee, Leskovec, Levent, Li,
  Li, Ma, Malik, Manning, Mirchandani, Mitchell, Munyikwa, Nair, Narayan,
  Narayanan, Newman, Nie, Niebles, Nilforoshan, Nyarko, Ogut, Orr,
  Papadimitriou, Park, Piech, Portelance, Potts, Raghunathan, Reich, Ren, Rong,
  Roohani, Ruiz, Ryan, Ré, Sadigh, Sagawa, Santhanam, Shih, Srinivasan,
  Tamkin, Taori, Thomas, Tramèr, Wang, Wang, Wu, Wu, Wu, Xie, Yasunaga, You,
  Zaharia, Zhang, Zhang, Zhang, Zhang, Zheng, Zhou, and
  Liang}]{foundation-models}
Rishi Bommasani, Drew~A. Hudson, Ehsan Adeli, Russ Altman, Simran Arora, Sydney
  von Arx, Michael~S. Bernstein, Jeannette Bohg, Antoine Bosselut, Emma
  Brunskill, Erik Brynjolfsson, Shyamal Buch, Dallas Card, Rodrigo Castellon,
  Niladri Chatterji, Annie Chen, Kathleen Creel, Jared~Quincy Davis, Dora
  Demszky, Chris Donahue, Moussa Doumbouya, Esin Durmus, Stefano Ermon, John
  Etchemendy, Kawin Ethayarajh, Li~Fei-Fei, Chelsea Finn, Trevor Gale, Lauren
  Gillespie, Karan Goel, Noah Goodman, Shelby Grossman, Neel Guha, Tatsunori
  Hashimoto, Peter Henderson, John Hewitt, Daniel~E. Ho, Jenny Hong, Kyle Hsu,
  Jing Huang, Thomas Icard, Saahil Jain, Dan Jurafsky, Pratyusha Kalluri,
  Siddharth Karamcheti, Geoff Keeling, Fereshte Khani, Omar Khattab, Pang~Wei
  Koh, Mark Krass, Ranjay Krishna, Rohith Kuditipudi, Ananya Kumar, Faisal
  Ladhak, Mina Lee, Tony Lee, Jure Leskovec, Isabelle Levent, Xiang~Lisa Li,
  Xuechen Li, Tengyu Ma, Ali Malik, Christopher~D. Manning, Suvir Mirchandani,
  Eric Mitchell, Zanele Munyikwa, Suraj Nair, Avanika Narayan, Deepak
  Narayanan, Ben Newman, Allen Nie, Juan~Carlos Niebles, Hamed Nilforoshan,
  Julian Nyarko, Giray Ogut, Laurel Orr, Isabel Papadimitriou, Joon~Sung Park,
  Chris Piech, Eva Portelance, Christopher Potts, Aditi Raghunathan, Rob Reich,
  Hongyu Ren, Frieda Rong, Yusuf Roohani, Camilo Ruiz, Jack Ryan, Christopher
  Ré, Dorsa Sadigh, Shiori Sagawa, Keshav Santhanam, Andy Shih, Krishnan
  Srinivasan, Alex Tamkin, Rohan Taori, Armin~W. Thomas, Florian Tramèr,
  Rose~E. Wang, William Wang, Bohan Wu, Jiajun Wu, Yuhuai Wu, Sang~Michael Xie,
  Michihiro Yasunaga, Jiaxuan You, Matei Zaharia, Michael Zhang, Tianyi Zhang,
  Xikun Zhang, Yuhui Zhang, Lucia Zheng, Kaitlyn Zhou, and Percy Liang. 2021.
\newblock \href {https://doi.org/10.48550/ARXIV.2108.07258} {On the
  opportunities and risks of foundation models}.

\bibitem[{Brown et~al.(2020)Brown, Mann, Ryder, Subbiah, Kaplan, Dhariwal,
  Neelakantan, Shyam, Sastry, Askell, Agarwal, Herbert-Voss, Krueger, Henighan,
  Child, Ramesh, Ziegler, Wu, Winter, Hesse, Chen, Sigler, Litwin, Gray, Chess,
  Clark, Berner, McCandlish, Radford, Sutskever, and Amodei}]{gpt3}
Tom Brown, Benjamin Mann, Nick Ryder, Melanie Subbiah, Jared~D Kaplan, Prafulla
  Dhariwal, Arvind Neelakantan, Pranav Shyam, Girish Sastry, Amanda Askell,
  Sandhini Agarwal, Ariel Herbert-Voss, Gretchen Krueger, Tom Henighan, Rewon
  Child, Aditya Ramesh, Daniel Ziegler, Jeffrey Wu, Clemens Winter, Chris
  Hesse, Mark Chen, Eric Sigler, Mateusz Litwin, Scott Gray, Benjamin Chess,
  Jack Clark, Christopher Berner, Sam McCandlish, Alec Radford, Ilya Sutskever,
  and Dario Amodei. 2020.
\newblock \href
  {https://proceedings.neurips.cc/paper/2020/file/1457c0d6bfcb4967418bfb8ac142f64a-Paper.pdf}
  {Language models are few-shot learners}.
\newblock In \emph{Advances in Neural Information Processing Systems},
  volume~33, pages 1877--1901. Curran Associates, Inc.

\bibitem[{Degen et~al.(2019)Degen, Hawkins, Graf, Kreiss, and
  Goodman}]{degen2019redundancy}
Judith Degen, Robert~XD Hawkins, Caroline Graf, Elisa Kreiss, and Noah~D
  Goodman. 2019.
\newblock When redundancy is rational: A {Bayesian} approach to
  ``overinformative’' referring expressions.
\newblock \emph{arXiv preprint arXiv:1903.08237}.

\bibitem[{Devlin et~al.(2019)Devlin, Chang, Lee, and
  Toutanova}]{devlin-etal-2019-bert}
Jacob Devlin, Ming-Wei Chang, Kenton Lee, and Kristina Toutanova. 2019.
\newblock \href {https://doi.org/10.18653/v1/N19-1423} {{BERT}: Pre-training of
  deep bidirectional transformers for language understanding}.
\newblock In \emph{Proceedings of the 2019 Conference of the North {A}merican
  Chapter of the Association for Computational Linguistics: Human Language
  Technologies, Volume 1 (Long and Short Papers)}, pages 4171--4186,
  Minneapolis, Minnesota. Association for Computational Linguistics.

\bibitem[{Firth(1957)}]{firth1957synopsis}
John~R Firth. 1957.
\newblock A synopsis of linguistic theory, 1930-1955.
\newblock \emph{Studies in linguistic analysis}.

\bibitem[{Goodman and Frank(2016)}]{goodman2016pragmatic}
Noah~D Goodman and Michael~C Frank. 2016.
\newblock Pragmatic language interpretation as probabilistic inference.
\newblock \emph{Trends in cognitive sciences}, 20(11):818--829.

\bibitem[{Goodman and Lassiter(2015)}]{goodman2015probabilistic}
Noah~D Goodman and Daniel Lassiter. 2015.
\newblock Probabilistic semantics and pragmatics: Uncertainty in language and
  thought.
\newblock \emph{The handbook of contemporary semantic theory, 2nd edition.
  Wiley-Blackwell}.

\bibitem[{Grice(1975)}]{grice1975logic}
Herbert~P Grice. 1975.
\newblock Logic and conversation.
\newblock In \emph{Speech acts}, pages 41--58. Brill.

\bibitem[{Heim and Kratzer(1998)}]{Heim1998}
Irene Heim and Angelika Kratzer. 1998.
\newblock \emph{Semantics in Generative Grammar}.
\newblock Blackwell.

\bibitem[{Heim(1982)}]{heim1982semantics}
Irene~Roswitha Heim. 1982.
\newblock \emph{The semantics of definite and indefinite noun phrases}.
\newblock University of Massachusetts Amherst.

\bibitem[{Kamp(1981)}]{kamp1981theory}
Hans~A Kamp. 1981.
\newblock A theory of truth and semantic representation formal methods in the
  study of language, part 1, ed. by jeroen groenendijk, theo janssen and martin
  and stokhof.
\newblock \emph{Amsterdam: Mathematisch Centrum}.

\bibitem[{Kratzer(2021)}]{sep-situations-semantics}
Angelika Kratzer. 2021.
\newblock {Situations in Natural Language Semantics}.
\newblock In Edward~N. Zalta, editor, \emph{The {Stanford} Encyclopedia of
  Philosophy}, {W}inter 2021 edition. Metaphysics Research Lab, Stanford
  University.

\bibitem[{Levinson et~al.(1983)Levinson, Levinson, and
  Levinson}]{levinson1983pragmatics}
Stephen~C Levinson, Stephen~C Levinson, and S~Levinson. 1983.
\newblock \emph{Pragmatics}.
\newblock Cambridge university press.

\bibitem[{Lewis(1979)}]{lewis1979scorekeeping}
David Lewis. 1979.
\newblock Scorekeeping in a language game.
\newblock In \emph{Semantics from different points of view}, pages 172--187.
  Springer.

\bibitem[{Li and Liu(2021)}]{li2021towards}
Shaojie Li and Yong Liu. 2021.
\newblock Towards sharper generalization bounds for structured prediction.
\newblock \emph{Advances in Neural Information Processing Systems}, 34.

\bibitem[{Lin et~al.(2022)Lin, Hilton, and Evans}]{truthfulqa}
Stephanie Lin, Jacob Hilton, and Owain Evans. 2022.
\newblock \href {https://doi.org/10.18653/v1/2022.acl-long.229}
  {{T}ruthful{QA}: Measuring how models mimic human falsehoods}.
\newblock In \emph{Proceedings of the 60th Annual Meeting of the Association
  for Computational Linguistics (Volume 1: Long Papers)}, pages 3214--3252,
  Dublin, Ireland. Association for Computational Linguistics.

\bibitem[{Merrill et~al.(2021)Merrill, Goldberg, Schwartz, and
  Smith}]{merrill2021provable}
William Merrill, Yoav Goldberg, Roy Schwartz, and Noah~A. Smith. 2021.
\newblock \href {https://doi.org/10.1162/tacl_a_00412} {{Provable Limitations
  of Acquiring Meaning from Ungrounded Form: What Will Future Language Models
  Understand?}}
\newblock \emph{Transactions of the Association for Computational Linguistics},
  9:1047--1060.

\bibitem[{Michael(2020)}]{blog2020}
Julian Michael. 2020.
\newblock \href
  {https://blog.julianmichael.org/2020/07/23/to-dissect-an-octopus.html} {To
  dissect an octopus: Making sense of the form/meaning debate}.

\bibitem[{Peters et~al.(2018)Peters, Neumann, Iyyer, Gardner, Clark, Lee, and
  Zettlemoyer}]{peters-etal-2018-deep}
Matthew~E. Peters, Mark Neumann, Mohit Iyyer, Matt Gardner, Christopher Clark,
  Kenton Lee, and Luke Zettlemoyer. 2018.
\newblock \href {https://doi.org/10.18653/v1/N18-1202} {Deep contextualized
  word representations}.
\newblock In \emph{Proceedings of the 2018 Conference of the North {A}merican
  Chapter of the Association for Computational Linguistics: Human Language
  Technologies, Volume 1 (Long Papers)}, pages 2227--2237, New Orleans,
  Louisiana. Association for Computational Linguistics.

\bibitem[{Potts(2020)}]{potts2020}
Christopher Potts. 2020.
\newblock \href
  {https://chrisgpotts.medium.com/is-it-possible-for-language-models-to-achieve-language-understanding-81df45082ee2}
  {Is it possible for language models to achieve understanding?}

\bibitem[{Wang et~al.(2013)Wang, Greiner, and Wang}]{wang2013consistency}
Shaojun Wang, Russell Greiner, and Shaomin Wang. 2013.
\newblock Consistency and generalization bounds for maximum entropy density
  estimation.
\newblock \emph{Entropy}, 15(12):5439--5463.

\end{thebibliography}
\bibliographystyle{acl_natbib}

\appendix

\onecolumn

\section{Limitations}

We derived a recipe for computing entailment in terms of text probabilities, hinting that entailment judgments may be decodable from LM predictions. Yet two key concerns qualify this conclusion.

\paragraph{Learnability}
We reduce entailment classification to computing probabilities in the \emph{target distribution} of an LM, not probabilities predicted by an LM.
In \autoref{sec:learnability}, we argue that the loss function of current LMs is not well suited to producing models from which entailment can be extracted.

\paragraph{Speaker Assumptions} 
Gricean speakers capture important factors influencing speech production in pragmatic theory, but human speakers are undoubtedly more complex. Based on \autoref{sec:generality}, we expect a similar isomorphism to hold under any reasonable speaker model, but the mathematical form may change and it may become harder to compute.

\section{Uncountable World Spaces} \label{sec:continuous}

In this section, we assume $\mathcal W$ is an uncountably infinite set with a a probability density function $p(w)$. We then define ``almost sure'' entailment as follows:

\begin{definition}
For $x, y \in \mathcal X$, we say $x$ almost surely entails $y$ (i.e., $\den{x} \sqsubseteq \den{y}$) if and only if
\begin{equation*}
    p( \den{x} \setminus \den{y} ) = 0 .
\end{equation*}
\end{definition}
\noindent Note that if $\mathcal W$ is countable, then $\mathcal A \sqsubseteq \mathcal B$ reduces to $\mathcal A \subseteq \mathcal B$. 
We can generalize \autoref{lem:sum-entails} as follows, which shows that all our results go through for almost sure entailment when $\mathcal W$ is uncountable.

\begin{lemma} \label{lem:sum-entails-cont}
Let $\mathbbm{1}_{\mathcal S}$ be the indicator function for set $\mathcal S$.
Let $f : \mathcal W \to \mathbb{R}$ be some function such that $\inf_{w \in \mathcal W} f(w) > 0$.
For any sets $\mathcal A, \mathcal B$ such that $\mathcal A \subseteq \mathcal B \subseteq \mathcal W$, then $p(\mathcal B \setminus \mathcal A) = 0$ if and only if
\begin{align*}
    \EX_{w \sim p(w)} \left[ \mathbbm{1}_{\mathcal A}(w) f(w) \right] &= \EX_{w \sim p(w)} \left[ \mathbbm{1}_{\mathcal B}(w) f(w) \right] .
\end{align*}
\end{lemma}
\begin{proof}
If $p(\mathcal B \setminus \mathcal A) = 0$, then the condition follows by construction.
We thus only need to show that the condition follows from $p(\mathcal B \setminus \mathcal A) = 0$.
Let $q = p(\mathcal B \setminus \mathcal A)$.
By linearity of expectation, we rewrite the premise condition as
\begin{align*}
    0
    &= \EX_{w \sim p(w)} \left[ \left(\mathbbm{1}_{\mathcal A}(w) - \mathbbm{1}_{\mathcal B}(w) \right) f(w) \right] \\
    &= \EX_{w \sim p(w)} \left[ \left(\mathbbm{1}_{\mathcal A}(w) - \mathbbm{1}_{\mathcal B}(w) \right) f(w) \mid w \in \mathcal B \setminus \mathcal A \right] q \\
    &+ \EX_{w \sim p(w)} \left[ \left(\mathbbm{1}_{\mathcal A}(w) - \mathbbm{1}_{\mathcal B}(w) \right) f(w) \mid w \not\in \mathcal B \setminus \mathcal A \right] (1 - q) \\
    &\geq \EX_{w \sim p(w)} \left[ f(w) \mid w \in \mathcal B \setminus \mathcal A \right] q .
\end{align*}
Letting $f^* = \inf_{w \in \mathcal W} f(w) > 0$, we get $0 \geq f^* q$. Since $f^* > 0$ and $q \geq 0$, $q = p(\mathcal B \setminus \mathcal A) = 0$. 
\end{proof}

\section{Gricean Speaker Proofs} \label{app:informative}

% Proof of \autoref{lem:cost}.
\lemCost*
\begin{proof}
Starting with the definition of an Gricean speaker, for any $x \in \mathcal X^*$ and $y \in \mathcal X \cup \{\omega\}$,
\begin{equation*}
    p(xy) = \EX_w \left[ p(x \mid w) p(y \mid x, w) \right] .
\end{equation*}
Now, letting $g(x, w) \triangleq p(x \mid w) / \left( \sum_{y'} \exp \left( \alpha I_\ell(y' \mid x; w) - c(y') \right) \right)$,
\begin{align*}
    p(xy) = \exp(-c(y)) \EX_w \left[ \exp(\alpha I_{\ell}(y \mid x; w)) g(x, w) \right] .
\end{align*}
We apply this identity to both sides of the fraction in the lemma statement:
\begin{align*}
    \frac{p(x\omega)}{p(xx)}
    &= \frac{\exp(-c(\omega)) \EX_w \left[ \exp( \alpha I_{\ell}(\omega \mid x; w)) g(x, w) \right] }{\exp(-c(x)) \EX_w \left[ \exp(\alpha I_{\ell}(x \mid x; w)) g(x, w) \right] } \\
    &= \frac{\exp(c(x))}{\exp(c(\omega))} \cdot \frac{ \EX_w \left[\exp(\alpha I_{\ell}(\omega \mid x; w)) g(x, w) \right] }{\EX_w \left[ \exp(\alpha I_{\ell}(x \mid x; w)) g(x, w) \right] } .
\end{align*}
Since $\den{x} \subseteq \den{\omega}$ and $\den{x} \subseteq \den{x}$, we know that the conditional information of both $\omega$ and $x$ given $x$ is $0$, and, thus,
\begin{equation*}
    \frac{p(x\omega)}{p(xx)}
    = \frac{\exp(c(x))}{\exp(c(\omega))} \cdot \frac{\EX_w \left[\exp(0) g(x, w) \right]}{\EX_w \left[\exp(0) g(x, w) \right]} \\
    = \frac{\exp(c(x))}{\exp(c(\omega))} .
\end{equation*}
\end{proof}

% Proof of \autoref{thm:informative}.
\thmInformative*
\begin{proof}
Recall from the proof of \autoref{lem:cost} that there exists a function $g(x, w)$ such that, for all $x \in \mathcal X^*$ and $y \in \mathcal X \cup \{\omega\}$,
\begin{equation*}
    p(xy) \propto \exp(-c(y)) \EX_w \left[ \exp(\alpha I_\ell(y \mid x; w)) g(x, w) \right] .
\end{equation*}
Thus, by \autoref{lem:cost}, the proposed distributional relation can be expanded as
\begin{align*}
    d_p(x, y)
    &\iff \frac{p(xy)}{p(x\omega)} = \frac{p(yy)}{p(y\omega)} \\
    &\iff p(xy) \cdot \frac{p(y\omega)}{p(yy)} = p(x\omega) \cdot \frac{p(xx)}{p(xx)} \\
    &\iff p(xy) \frac{\exp(c(y))}{\exp(c(\omega))} = p(xx) \frac{\exp(c(x))}{\exp(c(\omega))} \\
    &\iff p(xy) \exp(c(y)) = p(xx) \exp(c(x)) \\
    &\iff \EX_w \left[ \exp(\alpha I_\ell(y \mid x; w)) g(x, w) \right] = \EX_w \left[ \exp(\alpha I_\ell(x \mid x; w)) g(x, w) \right] .
\end{align*}
By \autoref{lem:sum-entails} \textcolor{\erratumcolor}{(Here is the error: \autoref{lem:sum-entails} does not apply! See \autoref{app:erratum})}, this holds if and only if, for all $w$,
\begin{align*}
    \exp(\alpha I_\ell(y \mid x; w)) &= \exp(\alpha I_\ell(x \mid x; w)) \\
    I_\ell(y \mid x; w) &= I_\ell(x \mid x; w) \\
    I_\ell(y \mid x; w) &= 0 \\
    \den{x}(w) \to \den{y}(w) &= 1 .
\end{align*}
We conclude the distributional relation holds if and only if $\den{x} \subseteq \den{y}$.
\end{proof}

\section{Proofs for Learning Bounds} \label{sec:proofs}

% Proof for \autoref{lem:log-concentration}.
\lemLogConcentration*
\begin{proof}
Without loss of generality, assume $p(z) > 0$.
With probabiliy $1 - (1 - p(z))^n$ over the draw of our sample, the random variable $\log \hat p(z)$ has finite variance defined by
\begin{equation*}
    \mathrm{Var} \left[ \log \hat p \right] = \frac{1}{n} \cdot \frac{1 - p(z)}{p(z)} \leq \frac{\mathfrak K_p(z)}{n} .
\end{equation*}
With finite variance, we can apply Chebyshev's inequality to conclude that
\begin{equation*}
    \mathrm{Pr} \left[  \abs{\log p(z) - \log \hat p(z)} \geq \epsilon \right] \leq \frac{ \mathrm{Var} \left[ \log \hat p \right] } {\epsilon^2} \leq \frac{ \mathfrak K_p(z) } {n \epsilon^2} .
\end{equation*}
Solving for $\delta \leq \mathrm{Pr} \left[  \abs{\log p(z) - \log \hat p(z)} \right]$, we get
\begin{align*}
    \delta &\leq \frac{ \mathfrak K_p(z) } {n \epsilon^2} \\
    \therefore \epsilon &\leq \sqrt{\frac{\mathfrak K_p(z)}{\delta n} } .
\end{align*}
We conclude that that with probability $1 - \delta - (1 - p(z))^n$,
\begin{equation*}
    \abs{\log p(z) - \log \hat p(z)} \leq \sqrt{\frac{\mathfrak K_p(z)}{\delta n} } .
\end{equation*}
\end{proof}

We now characterize the complexity factor $\mathfrak K_p(z)$ for uniformly truthful speakers.

\begin{lemma} \label{lem:uniform-log-complexity}
For all $z \in \mathcal X^* \cup \mathcal X^* \omega$ such that $\den{z}(p) > 0$, it holds that
\begin{equation*}
    \mathfrak K_p(z) \leq \frac{ \abs{\mathcal X}}{p(\den{z})} .
\end{equation*}
\end{lemma}
\begin{proof}
We start by deriving a lower bound on $p(z)$.
\begin{align*}
    p(z)
    &= \sum_w \frac{\den{z}(w)}{\sum_{z'} \den{z'}(w)} p(w) \\
    &\geq \sum_w \frac{\den{z}(w)}{\abs{\mathcal X}} p(w) \\
    &= \frac{\den{z}(p)}{\abs{\mathcal X}} .
\end{align*}
Applying this inequality to the definition of $\mathfrak K_p(z)$, we conclude that
\begin{equation*}
    \mathfrak K_p(z) \leq \frac{\abs{\mathcal X}}{\den{z}(p)} .
\end{equation*}
\end{proof}

% Proof for \autoref{thm:uniform-corpus}.

\autoref{lem:uniform-log-complexity} lets us to derive the following guarantee for estimating entailment scores using a corpus produced by uniformly truthful speakers:
\begin{restatable}{theorem}{thmUniformCorpus} \label{thm:uniform-corpus}
For a uniformly truthful speaker $p$,
let $u_p(x, y) = \log p(x\omega) - \log p(xy)$.
For $x,y \in \mathcal X$ such that $\den{xy}(p) > 0$ and $\delta > 0$, it holds with probability at least $1 - \delta - 2(1 - p(xy))^n$ that
\begin{equation*}
    \abs{ u_p(x, y) - u_{\hat p}(x, y) } \leq 2 \sqrt{ \frac{\abs{\mathcal X}}{p(\den{xy})} \cdot \frac{2}{\delta n} } .
\end{equation*}
\end{restatable}
\begin{proof}
We expand the difference in scores as follows:
\begin{equation*}
    \abs{u_p(x, y) - u_{\hat p}(x, y)}
    \leq \abs{\log p(x) - \log \hat p(x\omega)} + \abs{\log p(xy) - \log \hat p(xy)} .
\end{equation*}
We then apply \autoref{lem:log-concentration} with $\frac{\delta}{2}$. Since $p(x\omega) \geq p(xy)$, this implies that with probability $1 - \delta - 2(1 - p(xy))^n$,
\begin{align*}
    \abs{u_p(x, y) - u_{\hat p}(x, y)}
    &\leq \sqrt{\frac{2 \mathfrak K_p(x\omega)}{\delta n} } + \sqrt{\frac{2 \mathfrak K_p(xy)}{\delta n} } \\
    &\leq 2 \sqrt{\frac{2 \max \{ \mathfrak K_p(x\omega),\mathfrak K_p(xy) \}}{\delta n} } .
\end{align*}
Finally, we apply \autoref{lem:uniform-log-complexity} to conclude that
\begin{align*}
    \abs{ u_p(x, y) - u_{\hat p}(x, y) }
    &\leq 2 \sqrt{ \frac{\abs{\mathcal X}}{\min \{ \den{x\omega}(p), \den{xy}(p) \}} \cdot \frac{2}{\delta n} } \\
    &= 2 \sqrt{ \frac{\abs{\mathcal X}}{\den{xy}(p)} \cdot \frac{2}{\delta n} } .
\end{align*}
\end{proof}

We now characterize the complexity factor for Gricean speakers.

\begin{lemma} \label{lem:log-complexity}
Assume that $p(z \mid w)$ is a Gricean speaker with respect to listener $\ell$ and $\den{z}(w) = 1 \iff I_\ell(z ; w) \geq 0$. Then, for all $z \in \mathcal X^* \cup \mathcal X^*\omega$,
\begin{equation*}
    \mathfrak K_p(z) \leq \frac{\exp(c(z))}{p(\den{z})} .
\end{equation*}
\end{lemma}
\begin{proof}
We start by writing out the form of $p(z)$:
\begin{equation*}
    p(z) = \frac{\sum_w \exp(\alpha I_\ell(z; w)) p(w)}{\exp(c(z))} .
\end{equation*}
Because $z \in \mathcal X^* \cup \mathcal X^*\omega$, all terms where $\den{z}(w) = 1$ contribute at least $0$ information; other terms contribute negative information. Thus, we bound the information content of the ``true'' terms above $0$, and ignore the other terms to get the lower bound
\begin{align*}
    p(z)
    &\geq \frac{\sum_w \den{z}(w) \exp(0) p(w)}{\exp(c(z))} \\
    &= \frac{\sum_w \den{z}(w) p(w)}{\exp(c(z))} \\
    &= \frac{\den{z}(p)}{\exp(c(z))} .
\end{align*}
Plugging this into $\mathfrak K_p(z)$, we conclude that
\begin{equation*}
    \mathfrak K_p(z) \leq \frac{\exp(c(z))}{\den{z}(p)} .
\end{equation*}
\end{proof}

% Proof for \autoref{thm:log-bound}.
\thmLogBound*
\begin{proof}
We start by expanding $g_p(x, y)$:
\begin{align*}
    g_p(x, y)
    &= \log \frac{p(xy)}{p(x\omega)} - \log \frac{p(yy)}{p(y\omega)} \\
    &= \log p(xy) - \log p(x\omega) - \log p(yy) + \log p(y\omega) .
\end{align*}
Thus, following \autoref{thm:uniform-corpus}, we can bound
\begin{align*}
    \abs{g_p(x, y) - g_{\hat p}(x, y)} &\leq \abs{\log p(xy) - \log \hat p(xy)} + \abs{\log p(x\omega) - \log \hat p(x\omega)} \\
    &\quad + \abs{\log p(yy) - \log \hat p(yy)} + \abs{\log p(y\omega) - \log \hat p(y\omega)} .
\end{align*}
We apply \autoref{lem:log-concentration} to each term with $\frac{\delta}{4}$. Since $p(yy) \leq p(y\omega)$ and $p(xy) \leq p(x\omega)$, we get that with probability at least $1 - \delta - 4q^n$,
\begin{align*}
    \abs{g_p(x, y) - g_{\hat p}(x, y)}
    &\leq 4\sqrt{\frac{4 \max \{ \mathfrak K_p(xy), \mathfrak K_p(x\omega), \mathfrak K_p(yy), \mathfrak K_p(y\omega) \}}{\delta n}} \\
    &= 8\sqrt{\frac{\max \{ \mathfrak K_p(xy), \mathfrak K_p(x\omega), \mathfrak K_p(yy), \mathfrak K_p(y\omega) \}}{\delta n}} .
\end{align*}
Finally, we apply \autoref{lem:log-complexity} to conclude that, with probability at least $1 - \delta - 4q^n$,
\begin{align*}
    \abs{g_p(x, y) - g_{\hat p}(x, y)}
    &\leq 8\sqrt{ \max \left\{ \frac{\exp(c(xy))}{\den{xy}(p)}, \frac{\exp(c(x\omega))}{\den{x\omega}(p)}, \frac{\exp(c(yy))}{\den{yy}(p)}, \frac{\exp(c(y\omega))}{\den{y\omega}(p)} \right\} \cdot \frac{1}{\delta n}} \\
    &\leq 8\sqrt{ \frac{\exp( \max \{ c(xy), c(yy) \} )}{\den{xy}(p)} \cdot \frac{1}{\delta n}} .
\end{align*}
\end{proof}

We can use \autoref{cor:informative-no-cost} to derive a tighter version of \autoref{thm:log-bound} by removing the dependence on the uncommon string $yy$:
\begin{theorem} \label{cor:bound-cost}
Let $s_p(x, y) = \log \frac{p(x\omega)}{p(xy)} - c(y) + c(\omega)$ . Then, for all $x, y \in \mathcal X$ such that $\den{xy}(p) > 0$, for all $\delta > 0$, the following holds with probability $1 - \delta - 2(1-p(xy))^n$,
\begin{equation*}
    \abs{s_p(x, y) - s_{\hat p}(x, y)} \leq 2\sqrt{ \frac{\exp( c(xy) )}{p(\den{xy})} \cdot \frac{2}{\delta n}} .
\end{equation*}
\end{theorem}
The proof follows analogously to \autoref{thm:log-bound}.
The main improvement of \autoref{cor:bound-cost} compared to \autoref{thm:log-bound} is that the probability the bound holds no longer depends on the unlikely probability $p(yy)$. We also get the benefit that the cost complexity factor has been reduced to only depend on $c(xy)$ and obtain better constants ($2 \sqrt{2}$ instead of $8$), although these changes are likely less important than removing the dependence on $p(yy)$. Of course, the drawback is that we are assuming access to the cost function $c(y)$. If we have such access, though, the improvements in the bound suggest we may be able to extract entailment from a finite corpus of Gricean text with better sample complexity than if we did not.

\section{Sample Complexity Estimation Details} \label{app:estimation}

Assuming the approximation error in \autoref{thm:log-bound} is $\leq \epsilon$, we aim to solve the following inequality for $n$:
\begin{equation*}
    \epsilon \leq 8 \sqrt{ \frac{\exp(\max \{ c(xy), c(yy) \})}{p(\den{xy})} \cdot \frac{1}{\delta n} } .
\end{equation*}

\paragraph{Sentence Length} We make the simplifying assumption that $\max \{ c(xy), c(yy) \} = 2w(\ell + 1)$, where $\ell$ is a variable representing sentence length.\footnote{We write $\ell + 1$ instead of $\ell$ here for technical reasons: we want to guarantee that $q(z)$ can be a valid probability distribution.}
Let $\Sigma$ be the word-level vocabulary of English.
We estimate the value $w$ by assuming $q(z) = \exp(-w(\abs{z} + 1))$ is a valid prior over $\Sigma^*$ and solving for the unique value of $w$ to satisfy this condition:
\begin{align*}
    \sum_{z \in \Sigma^*} \exp(-w(\abs{z} + 1)) &= 1 \\
    \sum_{\ell = 0}^{\infty} \frac{\abs{\Sigma}^\ell}{\exp(w(\ell + 1))} &= 1 \\
    \exp(-w) \sum_{\ell = 0}^\infty \left( \frac{\abs{\Sigma}}{\exp(w)} \right)^\ell &= 1 \\
    % \frac{\exp(-w)}{1 - \frac{\abs{\Sigma}}{\exp(w)}} &= 1 \\
    % \exp(w) &= \abs{\Sigma} + 1 \\
    \therefore w &= \log(\abs{\Sigma} + 1) .
\end{align*}
This reveals that $w$ should be set $\geq 1$, but the question remains how to set $\abs{\Sigma}$.
In practice, we assume the speaker prior is defined over the support of all syntactically valid or likely strings in English, not over all possible strings as derived above. Letting $S$ be the word-level perplexity of English, we set $w$ according to
\begin{equation*}
    w \approx \log(S + 1) .
\end{equation*}
We set $S$ to the value estimated by GPT-3: $\sim 20$ nats/word \citep{gpt3}. Simplifying the numerator in the bound yields
\begin{align*}
    \exp(\log(21)(\ell + 1)) = 21^{\ell + 1} .
\end{align*}
Making the prior less strong, i.e., increasing $\abs{\Sigma}$ to be greater than this perplexity estimate, would only increase the number of samples needed to extract entailment judgments.
% Assuming $\Sigma$ (the word-level vocabulary of English) is $500$ suggests $w \approx 6$. Changing $\abs{\Sigma} = 1000000$ only increases $w$ by an order of magnitude: $w \approx 14$. For simplicity, we assume $w = 10$. Plugging this in simplifies the denominator of the complexity factor to $\exp(10(\ell + 1))$.

\paragraph{Truth Probability} We conservatively assume $p(\den{xy}) = \frac{1}{2}$, although in practice it may be smaller for more informative sentences. Reducing it would lead to higher sample complexity estimates.

\paragraph{Final Form} Putting together our estimates for sentence length and truth probability yields
\begin{align*}
    \epsilon &\leq 8 \sqrt{ \frac{2 \cdot 21^{\ell + 1}}{\delta n} } \\
    \therefore n &\leq \frac{128 \cdot 21^{\ell + 1}}{\delta \epsilon^2} .
\end{align*}
The final form captures the intuition that the likelihood of a string vanishes exponentially with its length, and that the base of this decay is roughly inversely proportional to the perplexity of the language.
In practice, we set $\delta = 0.1$ and $\epsilon = 1.0$. Changing the value of $\epsilon$ (the desired approximation accuracy) would shift the curve.

% \subsection{Perplexity-Based Bound}
% We compare the estimate from the last subsection against a simpler estimate that uses perplexity to estimate $p(z)$ directly.
% Overall, we do not find much of a difference between the two approaches.
% We assume that the probability of observing a sentence is related to this per-word perplexity and the sentence length according to $p(z) = (1/20)^{\ell + 1}$. Using this estimate as the complexity factor in the bound for \autoref{thm:log-bound} yields
% \begin{align*}
%     \epsilon &\leq 8 \sqrt{ \frac{20^{\ell + 1}}{\delta n} } \\
%     \therefore n &\leq \frac{64 \cdot 20^{\ell + 1}}{\delta \epsilon^2} .
% \end{align*}
% As with before, we assume $\epsilon, \delta = 0.1$ in our plot.

\section{General Relations and Speakers} \label{sec:existence}

So far, we have characterized concrete distributional relations that are isomorphic to entailment for different classes of speaker models. In this section, we analyze the conditions under which a distribution relation isomorphic to a semantic relation exists, given no assumptions about the speaker. Informally, we prove in \autoref{thm:existence} that a distributional isomorphism exists if and only if the speaker model depends on semantics ``at all''. This is a very weak condition, and should be satisfied by any reasonable model of natural speakers. Thus, we take this as evidence that any speaker model---not just the ones we have considered, admits a distributional relation isomorphic to entailment.

We now turn to the formal presentation of this result.
Let $M$ be the function that takes a set of worlds $\mathcal W$ and returns all semantic evaluation functions $\mu : \mathcal X \mapsto 2^{\mathcal W}$ over $\mathcal W$.
For a semantic evaluation function $\mu = \lambda x. \den{x}$, let $p_\mu$ be a speaker model parameterized by semantics $\mu$.

Say two semantic evaluation functions $\mu, \mu'$ are isomorphic with respect to $s$ if and only if, for all $x,y$,
\begin{equation*}
    S(\mu(x), \mu(y)) \iff S(\mu'(x), \mu'(y)) .
\end{equation*}

\begin{theorem} \label{thm:existence}
The following are equivalent for any speaker $p$ and semantic relation $s$:
\begin{enumerate}
    \item There exists a distribution relation $d$ such that, for all $\mathcal W$, for all $\mu \in M(\mathcal W)$, $s$ is isomorphic to $d_{p_\mu}$.
    \item For all $\mathcal W, \mathcal W'$, for all $\mu \in M(\mathcal W)$ and $\mu' \in M(\mathcal W')$ such that $\mu, \mu'$ are not isomorphic w.r.t. $s$, there exists $z \in \mathcal X^*$ such that $p_\mu(z) \neq p_{\mu'}(z)$.
\end{enumerate}
\end{theorem}

\begin{proof}
We will show that equivalence holds in both directions.

\paragraph{}\underline{Forward Direction:} We assume the second statement does not hold by way of modus tollens. Thus, there exists $\mathcal W, \mathcal W'$ with $\mu \in M(\mathcal W)$ and $\mu' \in M(\mathcal W')$ with $\mu, \mu'$ not isomorphic such that, for all $z \in \mathcal X^*$, $p_\mu(z) = p_{\mu'}(z)$. Thus, for all $d$ and sentences $x,y$,
\begin{align*}
    d_{p_\mu}(x, y) \iff d_{p_{\mu'}}(x, y) .
\end{align*}
But $\mu$ and $\mu'$ are not isomorphic, so there exist $x,y$ such that $S(\mu(x), \mu(y)) \notiff S(\mu'(x), \mu'(y))$. Thus, we can conclude that one of the following must hold:
\begin{align*}
    d_{p_\mu}(x, y) &\notiff S(\mu(x), \mu(y)) \\
    d_{p_{\mu'}}(x, y) &\notiff S(\mu'(x), \mu'(y)) .
\end{align*}
We conclude by modus tollens that the first statement implies the second.

\paragraph{}\underline{Backward Direction:} Assume the second statement holds. The function $f(\mu) = p_\mu$ is invertible up to isomorphism to $s$. In other words, there exists $g(p_\mu) = \mu^*$ such that, for all $x,y$,
\begin{equation*}
    S(\mu^*(x), \mu^*(y)) \iff S(\mu(x), \mu(y)) .
\end{equation*}
Then we define $d$ according to
\begin{align*}
    d_{p_\mu}(x, y)
        &\iff S(g(p_\mu)(x), g(p_\mu)(y)) \\
        &\iff S(\mu^*(x), \mu^*(y)) \\
        &\iff S(\mu(x), \mu(y)) .
\end{align*}
Thus, the second statement implies the first.
\end{proof}

\section{Experimental Details} \label{sec:details}

\subsection{Language Description}

\begin{figure}
    \centering
    \includegraphics[width=\textwidth]{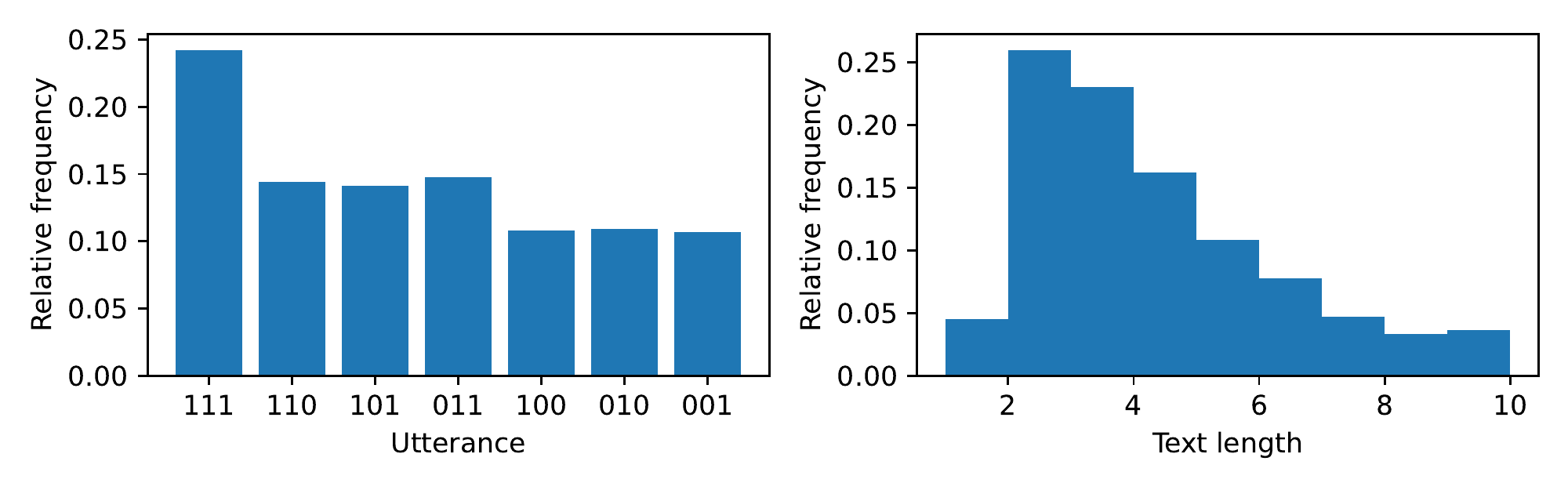}
    \caption{Properties of the data generated by the speaker in our experiments, with $\alpha=5$ and $c(x)=0.1\cdot\abs{x}$.}
    \label{fig:metadata}
\end{figure}

We set the vocabulary $\mathcal X = \{\texttt{100}, \texttt{010}, \texttt{001}, \texttt{110}, \texttt{011}, \texttt{111} \}$ and define $\mathcal W = \{1, 2, 3\}$. We refer to each three-digit binary string as an \emph{utterance}, and define the evaluation function for an utterance $x$ as $\den{x}(w) = 1 \iff x_w = \texttt{1}$. Thus, $\texttt{100}$ is true only in world $1$, while $\texttt{111}$ is true in all worlds (i.e., is tautological). We identify $\texttt{111}$ with the end of sequence $\omega$.

In line with our formal definitions, we define a \emph{text} $z$ as a concatenation of utterances $z_1\cdots z_n$ ending with $\omega$. Recall that we define the evaluation function over a text as the intersection of the evaluation functions of the utterances it contains. For our language, this reduces to
$\den{z}(w) = 1 \iff \forall i \left( z_i = \texttt{1} \right)$. Thus, $\texttt{011 101 111}$ is true only in $w_3$, and $\texttt{011 101 110 111}$ is true in no worlds (i.e., contradictory).

\subsection{Speaker Model Parameters}

We model the listener of the informative speaker as a literal listener \citep{goodman2016pragmatic}, which means our informative speaker is a rational speaker of depth $1$ in the language of rational speech acts.

We set $c(x) = 0.1 \cdot \abs{x}$, where $\abs{x}$ is the length of the string $x$. We set the rationality parameter $\alpha = 5$. These choices were made heuristically, by inspecting the the properties of the speaker's output, as summarized in Figure \ref{fig:metadata}. These parameters led to a relatively uniform distribution over utterances (except for the stop token \texttt{111} which is present in all texts), and a variety of text lengths without excessive redundancy. We found that larger values of $\alpha$ or of the coefficient for the cost function produced short texts, biasing maximally informative utterances (i.e., \texttt{100},  \texttt{010}, or \texttt{001}); while smaller values produced long, repetitive utterances or sometimes empty utterances.

\subsection{Training and Evaluation}

We sample a dataset from a speaker by independently sampling $n$ texts from the speaker model.
We generate datasets of varying size from each speaker, with the number of texts $n$ decreasing by factors of 2 from $10^7$ texts down to just 2 texts.

We train models of two kinds: a text frequency model, and a trigram model. The text frequency model simply assigns a probability to a text proportional to its frequency in the training data, assigning a small $\epsilon=10^{-20}$ probability to an unknown sequence. The trigram model is trained using NLTK's  \citep{bird-2006-nltk} MLE implementation, i.e., the probabilities are unsmoothed. We do not need to use smoothing due to the small number of possible trigrams in the language.

For evaluation data, we generate pairs of texts labeled for entailments. We include all pairs where each text is 6 utterances or shorter, except for utterances that are contradictory or consist only of the end of sequence token. The total number of test pairs is about 1.1M.

% While \autoref{thm:universal} says that 

% \subsection{Discussion}

% Informally, \autoref{thm:existence} says that if a speaker $p$ is sensitive to semantics at all, then \emph{any} semantic relation has a distributional relation that is isomorphic to it. Since the productions of natural speakers are generally assumed to be influenced by semantics in some way or another, this demonstrates that there is some distributional way of decoding semantics from practical LMs, no matter our specific theory of pragmatics.

% However, \autoref{thm:existence} is nonconstructive, and therefore does not reveal how we can \emph{decode} entailment from a generic LM. Furthermore, it does not guarantee that entailment can be resolved with a simple closed form as in the previous cases, or that it is efficiently computable or even decidable by a Turing machine. Thus, while \autoref{thm:existence} suggests that distributional signal contains full information about semantic relations like entailment, it is also important to establish whether it is easy to decode the semantic relation, i.e., $d_p$ is too computationally complex \citep{hewitt-etal-2021-conditional}. Thus, it is possible that data coming from more nuanced speakers---while technically encoding entailments according to \autoref{thm:existence}---could do so in a way that is too complex for any LM to ever recover. We believe this is an important question for future work to address in more detail.

\color{\erratumcolor}

\section{Erratum Derivation} \label{app:erratum}

Formally, $\den{x}$ can be partitioned into two sets of worlds:
\begin{compactenum}
    \item $Y = \den{x} \cap \den{y}$ where $x, y$ are both true
    \item $\tilde Y = \den{x} \setminus \den{y}$ where $x$ is true but $y$ is false
\end{compactenum}
Following the initial reasoning in the original proof of \autoref{thm:informative}, the entailment test is $0$ if and only if
\begin{align*}
    \EX_w [ \exp(\alpha I_\ell(y \mid x; w)) g(x, w) ] &= \EX_w [ \exp(\alpha \underbrace{I_\ell(x \mid x; w)}_{=0}) g(x, w) ] \\
    \EX_w [ \exp(\alpha I_\ell(y \mid x; w)) g(x, w) ] &= \EX_w [ g(x, w) ] .
\end{align*}
Flipping the equation for convenience, we get that
\begin{align*}
    \EX_w [ g(x, w) ] &= \EX_w [ \exp(\alpha I_\ell(y \mid x; w)) g(x, w) \mid w \in \den{x} ] \\
    &= p(Y) \EX_{w} [ \alpha \exp(I_\ell(y \mid x; w)) g(x, w) \mid w \in Y ] + \bcancel{p(\tilde Y) \EX_{w \in \tilde Y} [ \exp(\underbrace{I_\ell(y \mid x; w)}_{= - \infty}) \mid w \in \tilde Y ]} .
\end{align*}
Now, define
\begin{align*}
    I(Y) &\triangleq \EX_{w} [ \exp(I_\ell(y \mid x; w)) \mid w \in Y ] , \\
    I &\triangleq \EX_w [ g(x, w) ] .
\end{align*}
We conclude that the entailment test is $0$ if and only if $p(Y) I(Y) = I$.

Note that both $I(Y)$ and $I$ implicitly depend on $x$. They can be interpreted as measures of the conditional information of $y$ and the empty string after $x$, respectively.

\end{document}